\documentclass[lettersize,journal]{IEEEtran}
\usepackage{amsmath,amsfonts}
\usepackage{algorithm}
\usepackage{array}
\usepackage[caption=false,font=normalsize,labelfont=sf,textfont=sf]{subfig}
\usepackage{textcomp}
\usepackage{stfloats}
\usepackage{url}
\usepackage{verbatim}
\usepackage{graphicx}
\usepackage{cite}
\usepackage[flushleft]{threeparttable}
\usepackage{color}
\usepackage{xcolor}
\usepackage{multirow}
\usepackage{amssymb}
\usepackage{booktabs}
\usepackage{bbding}
\usepackage{pifont}
\usepackage{threeparttable}
\usepackage{bm}
\usepackage[breaklinks=true,colorlinks,bookmarks=false]{hyperref}
\usepackage{makecell}
\usepackage{multirow}
\usepackage{booktabs}
\usepackage{colortbl}  
\usepackage{algpseudocode}
\usepackage{xcolor}
\definecolor{hollywoodcerise}{rgb}{0.96, 0.0, 0.63}
\definecolor{lasallegreen}{rgb}{0.03, 0.47, 0.19}
\definecolor{hanpurple}{rgb}{0.32, 0.09, 0.98}
\definecolor{green(pigment)}{rgb}{0.0, 0.65, 0.31}

\definecolor{citecolor}{HTML}{229954}
\definecolor{best}{RGB}{225, 225, 225}
\definecolor{second}{RGB}{190, 205, 243}
\definecolor{aaa}{RGB}{232, 232, 232}
\definecolor{bbb}{RGB}{226, 240, 220}

\definecolor{alggray}{RGB}{232,232,232}
\definecolor{alggreen}{RGB}{205,255,235}
\definecolor{posblue}{RGB}{0,0,255}
\definecolor{negred}{RGB}{200,0,0}
\definecolor{actpink}{RGB}{255,235,242}

\begin{document}

\title{Just Noticeable Difference Modeling for Token \\ Compression  in 
Vision-Language-Action Models%
}

\author{
Zhuoyuan Li,~\IEEEmembership{Member,~IEEE,}
Rui Zhao,~\IEEEmembership{Member,~IEEE,}
Jin Wang,
Hanwei Zhu,~\IEEEmembership{Member,~IEEE,}\\
Cong Zhang,~\IEEEmembership{Member,~IEEE,}
Giuseppe Valenzise,~\IEEEmembership{Senior Member,~IEEE,}\\
Weisi Lin,~\IEEEmembership{Fellow,~IEEE,}
and Kin-Man Lam,~\IEEEmembership{Senior Member,~IEEE}\\\vspace{-1em}

\thanks{\textit{Corresponding author: Rui Zhao.}}
\thanks{Z. Li and K-M. Lam are with the Department of Electrical and Electronic Engineering, The Hong Kong Polytechnic University, Hong Kong
(e-mail: zhuoli@polyu.edu.hk, enkmlam@polyu.edu.hk)}
\thanks{R. Zhao, H. Zhu and W. Lin are with the School of Computer Science and Engineering, Nanyang Technological University, Singapore 639798 (e-mail: zhao.rui@ntu.edu.sg, hanwei.zhu@ntu.edu.sg, wslin@ntu.edu.sg)}
\thanks{Cong Zhang is with the School of Control Science and Engineering,
Shandong University (e-mail: congzhang@sdu.edu.cn)}

\thanks{Giuseppe Valenzise is with the CNRS, CentraleSupelec, Laboratoire des Signaux et Systèmes, Université Paris-Saclay, France (e-mail: giuseppe.valenzise@l2s.centralesupelec.fr).}
}

\markboth{Under Review}
{Li \MakeLowercase{\textit{et al.}}: A Sample Article Using IEEEtran.cls for IEEE Journals}

\maketitle
\begin{abstract}
Token compression has become a key technique for reducing the inference
cost of large foundation models, with approaches such as token pruning and KV-cache reuse widely adopted in vision-language models and recently explored for embodied agents. In embodied agents, tokens are no longer used only for perception and semantic understanding, but also directly support latency-sensitive closed-loop robot action prediction. Existing schemes typically guide token compression using redundancy or importance cues, such as visual similarity, attention scores, and saliency. However, these cues only indirectly reflect the quantity that
ultimately determines safe compression: how much a token can change before causing an unacceptable deviation in the downstream robot action. Such receiver-dependent tolerance is closely related to the core principle  of just noticeable difference (JND) theory.
Classical JND characterizes the tolerance to signal changes that remain imperceptible to the human visual system, while machine-oriented JND extends this tolerance boundary to downstream machine responses. Building on this progression, we introduce Action-JND, which extends JND modeling to embodied perception by defining noticeability through the language-conditioned action response of a vision-language-action (VLA) policy in closed-loop control, rather than through static downstream task responses in conventional machine perception. A visual-token change is therefore considered admissible only when the induced action deviation remains within a tolerated margin, rather than merely being imperceptible to humans or machines.
To instantiate this concept, we develop a lightweight token-wise JND estimator in deep visual-feature space that predicts the maximum tolerable perturbation while preserving the policy's action response. The resulting action-tolerance score serves as a plug-and-play ranking signal for representative VLA compression paradigms, including stale-KV reuse and token pruning, prioritizing action-tolerant tokens for compression while preserving action-sensitive ones. Experiments on the LIBERO robotic manipulation benchmark with OpenVLA and OpenVLA-OFT demonstrate that Action-JND consistently improves compression reliability, with particularly substantial gains under aggressive compression ratios.
\end{abstract}


\section{Introduction}
Large foundation models increasingly rely on long token sequences, making token compression a central challenge for efficient inference. In language models, long-context serving is often bottlenecked by the memory and bandwidth cost of the key-value (KV) cache, motivating KV-cache eviction, selection, and compression strategies that retain only a compact subset of informative historical tokens~\cite{zhang2023h2o,xiao2024efficient,li2024snapkv,cai2024pyramidkv}. In multi-modal large models, dense visual tokens further increase computational cost, and recent studies have explored visual token pruning, merging, and selection to reduce redundant computation while preserving downstream responses~\cite{bolya2022token,chen2024image,zhang2024sparsevlm,alvar2025divprune,yang2025visionzip,shang2025llava}. Meanwhile, feature coding provides an additional mechanism for compressing intermediate representations~\cite{gao2024dmofc,gao2025dt,huang2013feature,gao2025feature,li2026image,li2025embodied,gao2025compressed}. Similar efficiency challenges arise in vision-language-action (VLA) models for robotic manipulation, where robots must repeatedly process high-dimensional visual observations and generate actions in closed-loop control~\cite{zitkovich2023rt,kim2024openvla,kim2025fine,team2024octo}. Reducing inference cost is particularly important in this setting because
the policy operates within a closed-loop perception-action cycle, where lower latency enables more frequent action updates and more timely responses to environmental changes. Therefore, recent VLA acceleration methods have exploited spatial, semantic, and temporal redundancy through token pruning, layer pruning, dynamic scheduling, and adaptive cache reuse~\cite{xu2026vla,yang2026efficientvla,li2025sp,wang2025specprune,jiang2025better,pei2025action,liu2025vla,fang2025sqap}. In this paper, following common usage in efficient foundation model inference, we use the term token compression broadly to refer to reducing the effective token-processing cost, rather than to conventional bitrate-oriented source coding.

Despite their methodological differences, most token compression methods follow a common principle: they estimate compressibility from measures of redundancy or importance. A token is more likely to be pruned or reused if it is visually similar to neighboring or previous tokens, receives low attention, or exhibits limited saliency according to a specific scoring rule. These signals are useful, but they provide only indirect evidence regarding whether compression is truly safe. Redundancy indicates the extent to which information overlaps with other tokens, while importance estimates the degree to which the model attends to a token. Neither directly addresses the question that ultimately matters: \textbf{if a token is pruned or replaced by a stale representation, will the downstream behavior remain within an acceptable margin?}

\begin{figure*}[!t]
    \centering
    \vspace{-1.5em}
    \includegraphics[width=0.99\linewidth]{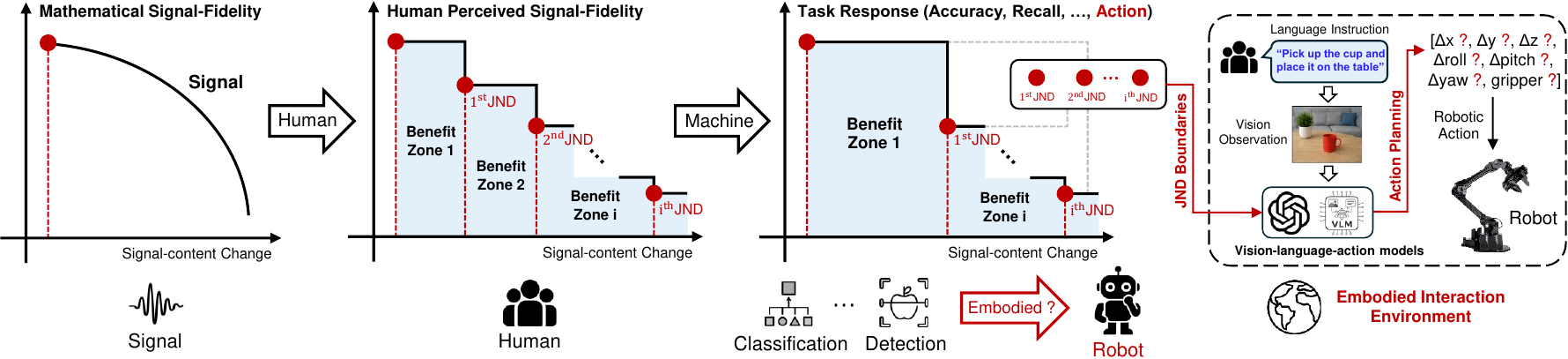}
    \vspace{-2mm}
    \caption{\textbf{From classical signal-oriented JND to machine-oriented JND for embodied perception.} (1) Classical signal coding characterizes  the trade-off between mathematical signal fidelity and signal-content change, where compression quality is usually evaluated using signal-level distortion. (2) Human-oriented JND introduces perceptual benefit zones, where signal changes remain unnoticeable to human observers despite measurable signal distortion. (3) Machine-oriented JND extends receiver-dependent tolerance to machine responses, where noticeability is determined by downstream task responses, such as classification and detection. (4) For embodied perception, we further instantiate the machine-oriented JND boundary with respect to action responses in a VLA-based embodied interaction environment, where noticeability is determined by whether changes in visual representations induce a tolerable deviation in the language-conditioned robot action during closed-loop control.}
\label{fig:intro_jnd_concept}
    \vspace{-1.5em}
\end{figure*}

This question is closely related to a fundamental principle of compression: a representation can be altered as long as the resulting change remains acceptable to its intended receiver. In conventional image/video coding, this principle is usually expressed through the rate-distortion (R-D) trade-off, where lower bitrates are achieved at the expense of increased mathematical signal distortion~\cite{sullivan1998rate,bross2021overview,sullivan2012overview,li2025ustc}, as shown in the leftmost part of Fig.~\ref{fig:intro_jnd_concept}. However, mathematical distortion alone does not necessarily indicate whether a compressed signal is perceptually degraded. Perceptual coding incorporates the human receiver into this trade-off, e.g., leveraging the just noticeable difference (JND) characteristics of the human visual system (HVS)~\cite{chou1995perceptually,jayant2002signal,lin2021progress,wu2013just}, as shown in the second  part of Fig.~\ref{fig:intro_jnd_concept}. Within the JND range, a signal may differ numerically from its original version yet remain visually indistinguishable to human observers, forming a benefit zone for compression. From this perspective, JND can be interpreted as a receiver-aware tolerance boundary that specifies the maximum admissible representation change without causing a perceptible variation.

This receiver-aware interpretation becomes more general when the compressed representation is consumed by a machine model rather than directly observed by humans. The compressed entity may take the form of a visual token, an intermediate feature representation, or a multimodal embedding, while the receiver is the downstream model that processes it. In such cases, the notion of “noticeability” should be defined by changes in the receiver’s output response. 
As shown in the third part of Fig.~\ref{fig:intro_jnd_concept}, machine-oriented JND extends the benefit-zone concept to machine receivers, where the boundary is characterized by downstream task responses, such as classification accuracy, detection recall, or machine satisfaction ratios~\cite{jin2021just,zhang2021just,zhang2024perceptual,shen2025transferring,gao2021towards}. Recent studies have further expanded this idea to large multimodal models and deep visual features, where noticeability is quantified through multimodal response variation or task-preserving feature perturbations~\cite{chen2025just,zhao2026just}. These studies suggest that JND is not restricted to HVS but provides a general way for quantifying how much representation can change while preserving the response of a designated receiver.

Following this progression, embodied intelligence  provides a new machine-response perspective for representation compression, as shown in the rightmost part of Fig.~\ref{fig:intro_jnd_concept}. In this setting, the machine receiver of compressed visual tokens is a language-conditioned robotic policy operating in a closed-loop control environment. Unlike conventional machine-perception tasks whose responses are typically class labels, bounding boxes, or textual outputs, the response of a robotic policy is a robot action, comprising end-effector translation $(\Delta x, \Delta y, \Delta z)$, rotation $(\Delta \mathrm{roll}, \Delta \mathrm{pitch}, \Delta \mathrm{yaw})$, and gripper control $(g)$. Therefore, the quality of token compression should be evaluated according to whether it induces action-level deviations, alters action planning, or degrades task execution, rather than according to image fidelity or generic task performance. Motivated by this observation, we formulate an action-conditioned tolerance boundary, termed \textbf{Action-JND}: the maximum compression-induced token change that preserves the robotic policy’s
action prediction within a tolerated deviation. Compared with existing machine-oriented JND formulations that mainly consider static downstream task responses~\cite{jin2021just,zhang2021just,zhang2024perceptual,zhao2026just}, Action-JND redefines the noticeability to robot actions in closed-loop control. Specifically, the receiver is instantiated as a VLA policy, the tolerated change is measured through action deviation, and the considered representation changes arise from token compression during VLA inference.

This action-level perspective provides a principled basis for re-examining existing VLA acceleration schemes in robotic control. During VLA inference, pruning a visual token or reusing a stale KV representation should be considered safe only when the resulting representation change remains within the Action-JND boundary. However, current token-level VLA acceleration methods usually approximate this safety criterion using indirect proxy signals. For example, VLA-Cache selects reusable tokens according to visual stability and attention-based risk estimates~\cite{xu2026vla}; VLA-Pruner removes visual tokens based on temporal-aware semantic and action attention~\cite{liu2025vla}; and EfficientVLA combines visual-token selection, layer pruning, and temporal feature caching to exploit redundancy throughout the VLA pipeline~\cite{yang2026efficientvla}. Recent methods have further explored model scheduling, self-speculative pruning, differentiable token pruning, and action-aware dynamic pruning~\cite{li2025sp,wang2025specprune,jiang2025better,pei2025action}. These approaches effectively reduce computational cost by estimating token stability, importance, or redundancy, but they do not explicitly quantify the action tolerance of each token. This limitation becomes particularly pronounced under aggressive pruning or cache reuse. For instance, a visually stable token located near the gripper, contact region, or object boundary may be highly influential for action generation and therefore unsuitable for compression. Conversely, a semantically salient token may admit substantial compression without affecting
the predicted action. These observations suggest that token compression in VLA models should be guided not only by whether a token appears important or redundant, but also by how much its representation can change while preserving the policy’s action response.

In this paper, we extend receiver-dependent JND modeling tailored to embodied perception through an action-conditioned tolerance boundary for VLA models. Unlike existing machine-oriented JND formulations that mainly consider static downstream task responses, Action-JND determines whether a token representation change is noticeable through the induced robot-action deviation in closed-loop control. To estimate this tolerance  boundary, we instantiate Action-JND in the deep visual-feature space and develop a lightweight token-wise JND estimator on VLA models. During training, the JND estimator searches for token perturbations that are as large as possible while keeping the robotic policy's action prediction within an allowed deviation. The learned perturbation magnitude serves as an action-tolerance score, where high-score tokens are more action-tolerant and low-score tokens are more action-sensitive. Based on this score, we further develop two representative VLA acceleration paradigms: KV-cache reuse and token pruning. For KV-cache reuse, it ranks candidate tokens for stale-KV reuse according to their estimated action tolerance. For token pruning, Action-JND prioritizes the removal of highly action-tolerant tokens under a prescribed pruning budget. In both cases, the compression objective is shifted from preserving proxy cues to directly preserving downstream robot actions under compression. Extensive experiments across different compression ratios demonstrate that the proposed approach  achieves superior performance over existing token-pruning and KV-cache-based VLA acceleration methods, with particularly substantial gains in aggressive compression regimes, resulting in more reliable acceleration and stronger action preservation.

The contributions of this paper are summarized as follows:
\begin{itemize}
\item We extend classical JND theory toward embodied perception, termed Action-JND, and formulate token compression in VLA models as an action-conditioned tolerance modeling problem, where noticeability is determined by the induced robot-action deviation.

\item We instantiate the Action-JND in deep visual-feature space with a token-wise JND estimator, which learns tolerable token perturbations and converts
their magnitudes into a criterion for token compression in VLA models.

\item We integrate the learned Action-JND criterion into two representative VLA token compression paradigms, including KV cache reuse and token pruning, enabling more reliable stale-KV reuse and token removal.

\item Extensive experiments across a broad range of compression ratios demonstrate that Action-JND improves compression reliability over existing caching and   pruning schemes for VLA models.
\end{itemize}

\section{Related Work}
In this section, we review three research areas related to the proposed framework. First, we introduce recent VLA models and their action-generation paradigms. Then, we discuss efficient inference and token compression methods for large models, especially token pruning and temporal KV cache reuse in VLA models. Finally, we review JND modeling from human perception to machine-oriented tolerance estimation, which motivates our action-aware JND formulation for embodied perception.

\subsection{Vision-Language-Action Models}
VLA models extend vision-language foundation models from passive perception and language understanding to active robotic control. Instead of producing only textual responses, VLA models map visual observations and language instructions to executable robot actions. Early representative systems such as RT-2 formulate robot control as a vision-language generation problem by representing actions as language-like tokens and co-fine-tuning vision-language models with robotic trajectory data~\cite{zitkovich2023rt}. OpenVLA further advances this paradigm through an open-source 7B-parameter VLA model trained on large-scale robot demonstrations, where continuous robot actions are discretized into action tokens and generated  autoregressively~\cite{kim2024openvla}. Other generalist robot policies and VLA-style models, such as Octo and $\pi_0$, explore different policy architectures and action representations to enable scalable robot control~\cite{team2024octo,black2024pi_0}. Collectively, these studies show the potential of large pretrained multimodal  models for language-conditioned robotic manipulation, while simultaneously exposing the substantial inference cost associated with large visual encoders and language backbones.

More recent efforts have focused on improving the efficiency and deployment
practicality of VLA models by redesigning the action-generation process itself.
OpenVLA-OFT studies several key fine-tuning choices, including autoregressive
versus parallel decoding, discrete versus continuous action representations,
and next-token prediction versus regression or diffusion-based objectives
~\cite{kim2025fine}. By combining parallel decoding, action
chunking, continuous action prediction, and an $\ell_1$ regression
objective, OpenVLA-OFT improves both policy quality and action generation
speed. These developments indicate that VLA models may adopt different
action parameterizations, ranging from autoregressive discrete action
tokens to continuous action chunks. Despite these advances, current VLA
policies still rely on large visual encoders and language backbones and
repeatedly process dense visual tokens during closed-loop control.
Therefore, reducing inference cost while maintaining reliable action
generation remains essential for practical VLA deployment, motivating
the development of efficient inference and token-compression techniques.

\subsection{Efficient Inference and Token Compression for Large Model}

Token compression has been widely studied for efficient foundation-model inference. In large language models, KV-cache eviction, selection, and compression reduce the memory and bandwidth cost of long-context inference~\cite{zhang2023h2o,xiao2024efficient,li2024snapkv,cai2024pyramidkv}, while vision transformers and vision-language models reduce dense visual-token computation through token merging, pruning, and selection~\cite{bolya2022token,chen2024image,zhang2024sparsevlm,alvar2025divprune,yang2025visionzip,shang2025llava}. However, these methods are primarily developed for generic language, vision,
or multimodal inference, and their extension to embodied robotic control remains comparatively underexplored.

When token compression is applied to embodied agents, additional requirements emerge. VLA models and action policies operate within closed-loop robotic control systems and repeatedly process temporally adjacent observations to predict low-level actions under strict latency constraints. Since inference latency directly limits how frequently the policy can update robot actions, efficient inference is critical for maintaining responsive closed-loop control. Therefore, token compression must not only exploit spatial and temporal redundancy, but also preserve the visual information required for language-conditioned perception and action generation. Existing VLA acceleration methods can be broadly categorized into two groups. The first group improves model architectures or action-generation mechanisms, including parallel decoding, action chunking, quantization-aware pruning, and lightweight policy design~\cite{kim2025fine,fang2025sqap}. These approaches reduce computation by compacting the policy architecture, action-generation pipelines, or model parameters. However, dense visual-token representations are generally propagated through most of the VLA backbone, leaving a substantial amount of redundancy in intermediate visual representations unexploited. The second group performs token-level acceleration by exploiting redundancy within visual representations~\cite{xu2026vla,yang2026efficientvla,li2025sp,wang2025specprune,jiang2025better,pei2025action,liu2025vla}. For example, VLA-Cache identifies visually stable tokens across consecutive observations and reuses their cached KV states, while using decoder attention to protect task-relevant regions~\cite{xu2026vla}. VLA-Pruner jointly considers semantic understanding and action execution by combining vision-language prefill attention with temporally smoothed action-decoding attention for token pruning~\cite{liu2025vla}. Other methods introduce model scheduling, self-speculative pruning, differentiable token pruning, action-aware dynamic pruning, and  quantization-aware pruning to improve VLA efficiency~\cite{li2025sp,wang2025specprune,jiang2025better,pei2025action,fang2025sqap}.

Although these approaches achieve promising acceleration performance, their compression decisions remain largely driven by proxy signals, such as visual stability, attention-based importance, scheduling confidence, or token redundancy. While such cues are useful for identifying potentially compressible tokens, they do not explicitly estimate the action sensitivity of individual tokens, namely the extent to which a token can be altered before the predicted robot action is affected. This limitation is particularly important in robotic manipulation scenarios, where visually stable or weakly attended regions may nevertheless play a critical role in action generation. Therefore, VLA token compression requires a criterion that directly evaluates whether a token can be pruned or reused while maintaining the resulting action deviation within an acceptable range. This motivates the proposed Action-JND formulation, which reframes token compression as an action-tolerance modeling problem.

\vspace{-2mm}
\subsection{JND Modeling}
JND was originally introduced in human perception to characterize the minimum signal change that becomes perceptible to human observers. In image and video processing, JND has been widely used in perceptual coding, where distortions below the human JND threshold are mathematically measurable but visually unnoticeable, forming a benefit zone for compression, watermarking, quality assessment, and resource allocation~\cite{chou1995perceptually,jayant2002signal,lin2021progress}. Classical visual JND models mainly exploit characteristics of HVS, including luminance adaptation and contrast sensitivity. Subsequent studies further construct subjective JND datasets and develop learning-based predictors for picture-wise JND assessment, JND-critical region localization, and JND map estimation~\cite{lin2022large,chen2023localization,jiang2022toward,wang2024metajnd}. These works establish a fundamental principle of JND modeling: safe distortion should be determined by the receiver’s response, rather than by signal fidelity alone.

This receiver-dependent perspective has gradually been extended from human perception to machine perception. In deep machine vision, DMV-JND shows that images exhibiting large signal distortion can still be tolerable for classification models~\cite{jin2021just}. For machine-oriented visual coding, the concepts of just recognizable distortion and satisfied machine ratio  further characterize whether compressed visual signals preserve downstream machine-task performance~\cite{zhang2021just,zhang2024perceptual}. More recently, JND-based tolerance modeling has been extended to large multimodal models and deep visual features. LMM-JND investigates the minimum visual change that can be perceived by large multimodal models and reflected in their textual responses~\cite{chen2025just}, whereas FeatJND estimates the maximum tolerable feature perturbation that can be introduced while preserving downstream classification, detection, and instance segmentation performance~\cite{zhao2026just}. These studies show that JND is no longer limited to human perception, but can serve as a general tolerance modeling principle for machine receivers.

However, existing machine-oriented JND formulations mainly define
noticeability through recognition outputs, multimodal responses, or static downstream task performance, while JND modeling for embodied perception remains largely unexplored. In embodied perception, the machine receiver interacts with the environment and produces robot actions conditioned on both visual observations and language instructions. Therefore, the tolerance of a visual representation change
should be determined by its induced action deviation rather than by signal distortion or generic task accuracy. Motivated by this observation, we introduce Action-JND, which further extends receiver-dependent JND modeling to embodied perception through action-conditioned noticeability. This perspective further provides a receiver-aware criterion for reliable token compression in robotic manipulation.

\section{JND Modeling for Embodied Perception}
\subsection{Preliminaries}
JND characterizes the minimum signal change that becomes noticeable to a receiver. Equivalently, it can be interpreted as the maximum distortion that can be introduced while remaining unnoticeable under a given response criterion. In classical visual perception, the receiver is the HVS, and a signal change is considered imperceptible if it does not cause noticeable perceptual degradation~\cite{chou1995perceptually,jayant2002signal,lin2021progress}. Therefore, JND defines a receiver-dependent tolerance boundary: a distortion is regarded as acceptable not because it is numerically small, but because it does not alter the receiver’s perception. This receiver-dependent principle has subsequently been extended from human
perception to machine perception, where noticeability is determined by
changes in downstream machine responses. Existing machine-oriented JND
studies examine whether perturbations applied to visual signals preserve recognition performance or machine satisfaction
ratios~\cite{jin2021just,zhang2021just,zhang2024perceptual}. More recently, FeatJND further extends this principle from input  signals to intermediate feature representations~\cite{zhao2026just}. Given an intermediate feature tensor $f$ and a downstream task head $h(\cdot)$, FeatJND estimates a perturbation map $\delta$ that is as large as possible while preserving
the downstream task response. This formulation shifts the tolerance criterion from input-signal changes to task-preserving feature
perturbations, providing a basis for modeling JND over token representations.

Building upon this progression, we further extend JND modeling toward embodied
perception. In VLA-based robotic manipulation, the machine receiver is
instantiated as a robotic policy that consumes visual representations
together with language instructions to generate control actions
~\cite{zitkovich2023rt,kim2024openvla,kim2025fine,team2024octo}. The receiver response to be preserved is therefore the language-conditioned robot action produced by the policy. Accordingly, a change in a visual-token representation becomes noticeable if the predicted action deviates beyond an acceptable tolerance level. We therefore define Action-JND as the maximum visual-token perturbation that can be introduced while preserving the action response of a VLA policy.

\subsection{Receiver-Dependent Formulation}
To formulate the above intuition, we first express JND in a general receiver-dependent form. Let $x$ denote an input representation and $\Delta$ denote a signal or feature perturbation. For a receiver $\mathcal{R}$, let $r_{\mathcal{R}}(x)$ denote its response to $x$, and let $D_{\mathcal{R}}(\cdot,\cdot)$ measure the discrepancy  between two responses. The specific form of $D_{\mathcal{R}}$ depends on the receiver and its response space. For example, it may represent perceptual difference for human observers, task discrepancy for machine perception models, or an action discrepancy for robotic policies. A receiver-dependent JND can be formulated as,
\begin{equation}
\Delta_{\mathcal{R}}^{\star}
= \arg\max_{\Delta} \; M(\Delta),
\quad \mathrm{s.t.}\,\,\,
D_{\mathcal{R}}\big(r_{\mathcal{R}}(x), r_{\mathcal{R}}(x+\Delta)\big)
\le \epsilon_{\mathcal{R}}.
\label{eq:receiver_jnd}
\end{equation}
where $M(\Delta)$ measures the perturbation magnitude, such as $\ell_1$ or $\ell_2$ norm. $\epsilon_{\mathcal{R}}$ is the tolerated response variation. 
Here, $M(\cdot)$ and $D_{\mathcal{R}}(\cdot,\cdot)$ are scalar-valued functions, $\epsilon_{\mathcal{R}}$ is a scalar tolerance threshold, and $x$ and $\Delta$ are vectors or tensors. For human-oriented JND, $\mathcal{R}$ is the HVS, and the constraint requires perceptual indistinguishability.

For machine-oriented JND, the receiver becomes a downstream machine task~\cite{zhao2026just}. Let $f$ denote an intermediate feature tensor and $h(\cdot)$ denote the downstream task head. The feature-level JND perturbation can be formulated as,
\begin{equation}
\delta^{\star}
= \arg\max_{\delta} \; M(\delta),
\quad \mathrm{s.t.} \,\,\,
D_{\mathrm{task}}\big(h(f),h(f+\delta)\big)
\le \epsilon_{\mathrm{task}},
\label{eq:feature_jnd}
\end{equation}
where $D_{\mathrm{task}}(\cdot,\cdot)$ measures the downstream task discrepancy, and $\epsilon_{\mathrm{task}}$ denotes the tolerated task variation. Depending on application, $D_{\mathrm{task}}$ can be instantiated by different forms, such as classification loss, regression error, distribution divergence, or other task-specific measures. Eq.~\eqref{eq:feature_jnd} instantiates the general receiver-dependent formulation in Eq.~\eqref{eq:receiver_jnd} for machine perception, where the receiver is a downstream task head and the response criterion is defined by task-response preservation.

\subsection{Action-JND Formulation}
For embodied perception, we instantiate the machine receiver as a VLA policy operating in a robotic system. At timestep $t$, let
$F_t =
[f_{t,1}, f_{t,2}, \ldots, f_{t,N}]^{\top}
\in \mathbb{R}^{N\times D}$
denote the visual-token features, where $N$ is the number of visual tokens and $D$ is the feature dimension. Let $\ell$ denote the language instruction. A frozen VLA policy $\pi_{\theta}$ predicts the robot action as $a_t = \pi_{\theta}(F_t,\ell)$. 
For the robotic manipulation setting considered in this work,
continuous-action policies can represent the action by end-effector
translation, rotation, and gripper control~\cite{kim2024openvla,kim2025fine}:
$a_t =
[\Delta x_t,\Delta y_t,\Delta z_t,
\Delta \mathrm{roll}_t,\Delta \mathrm{pitch}_t,
\Delta \mathrm{yaw}_t,g_t]$. Here, $a_t$ is a vector whose components are scalars. For discrete-action generation policies, the response can instead be represented
by action-token distributions~\cite{kim2024openvla}.
The Action-JND formulation is not restricted to this action space and can be extended to other embodied agents by defining the policy response and action discrepancy according to their physical action spaces. For simplicity, we use a single-step action in the formulation, while the same response-level formulation applies to multi-step action chunks, as instantiated by OpenVLA and OpenVLA-OFT, respectively. Let $\tilde{a}_t = \pi_{\theta}(F_t + \Delta_t, \ell)$ denote the action predicted from perturbed features. 
We define action-level noticeability according to the discrepancy between the clean and perturbed policy responses, 
$D_{\mathrm{act}}\big(
\mathcal{R}_{\theta}(F_t,\ell),
\mathcal{R}_{\theta}(F_t+\Delta_t,\ell)
\big)$.
Following the same JND principle~\cite{zhao2026just}, Action-JND is defined as the maximum visual-token perturbation that remains unnoticeable to VLA policy:
\begin{equation}
\begin{split}
& \Delta_t^{\star}
= \arg\max_{\Delta_t} \; M(\Delta_t), \\
& \mathrm{s.t.}\quad
D_{\mathrm{act}}
\big(
\mathcal{R}_{\theta}(F_t,\ell),
\mathcal{R}_{\theta}(F_t+\Delta_t,\ell)
\big)
\le \epsilon_a ,
\end{split}
\label{eq:action_jnd_basic}
\end{equation}
where $\mathcal{R}_{\theta}(\cdot)$ denotes the policy response to be preserved. For discrete-action policies~\cite{kim2024openvla}, it is the action-token distribution $P_\theta(a \mid F_t, \ell)$; for continuous-action policies~\cite{kim2025fine}, it is the regressed action vector $a_t = \pi_\theta(F_t, \ell)$.
$\epsilon_a$ denotes the tolerated action deviation. Compared with feature-level JND in Eq.~\eqref{eq:feature_jnd},
Action-JND follows the same receiver-dependent JND principle but shifts
the response criterion toward embodied action generation: the machine
receiver is instantiated as a VLA policy rather than a task head, and the response is a robot action in closed-loop control rather than a classification, detection, segmentation, or textual output.

The action discrepancy $D_{\mathrm{act}}$ is instantiated according to how VLA policy parameterizes its action output. 
For discrete-action policies such as OpenVLA~\cite{kim2024openvla}, the response is represented by the action-token distribution $P_\theta(a\mid F_t,\ell)$, and $D_{\mathrm{act}}$ is its Kullback-Leibler (KL) divergence under perturbation:
\begin{equation}
D_{\mathrm{act}}
=
D_{\mathrm{KL}}
\Big(
P_{\theta}(a|F_t,\ell)
\; \big\Vert\;
P_{\theta}(a|F_t+\Delta_t,\ell)
\Big).
\label{eq:discrete_action_distance}
\end{equation}
For continuous-action policies that directly regress action chunks, such as OpenVLA-OFT~\cite{kim2025fine}, the action instead lies in a continuous space, and $D_{\mathrm{act}}$ is computed as a weighted $\ell_1$ regression distance:
\begin{equation}\small
D_{\mathrm{act}}(a_t,\tilde{a}_t)
= \lambda_p \|\Delta p_t-\Delta \tilde{p}_t\|_1
+ \lambda_r \|\Delta r_t-\Delta \tilde{r}_t\|_1
+ \lambda_g |g_t-\tilde{g}_t|.
\label{eq:continuous_action_distance}
\end{equation}
where $\Delta p_t$ denotes the end-effector translation, $\Delta r_t$ denotes the end-effector rotation, and $g_t$ denotes the gripper command. $\lambda_p$, $\lambda_r$, and $\lambda_g$ weight the translation, rotation, and gripper-control discrepancies, respectively.

\begin{figure*}[!t]
    \centering    
    \vspace*{-15pt}
    \includegraphics[width=0.95\linewidth]{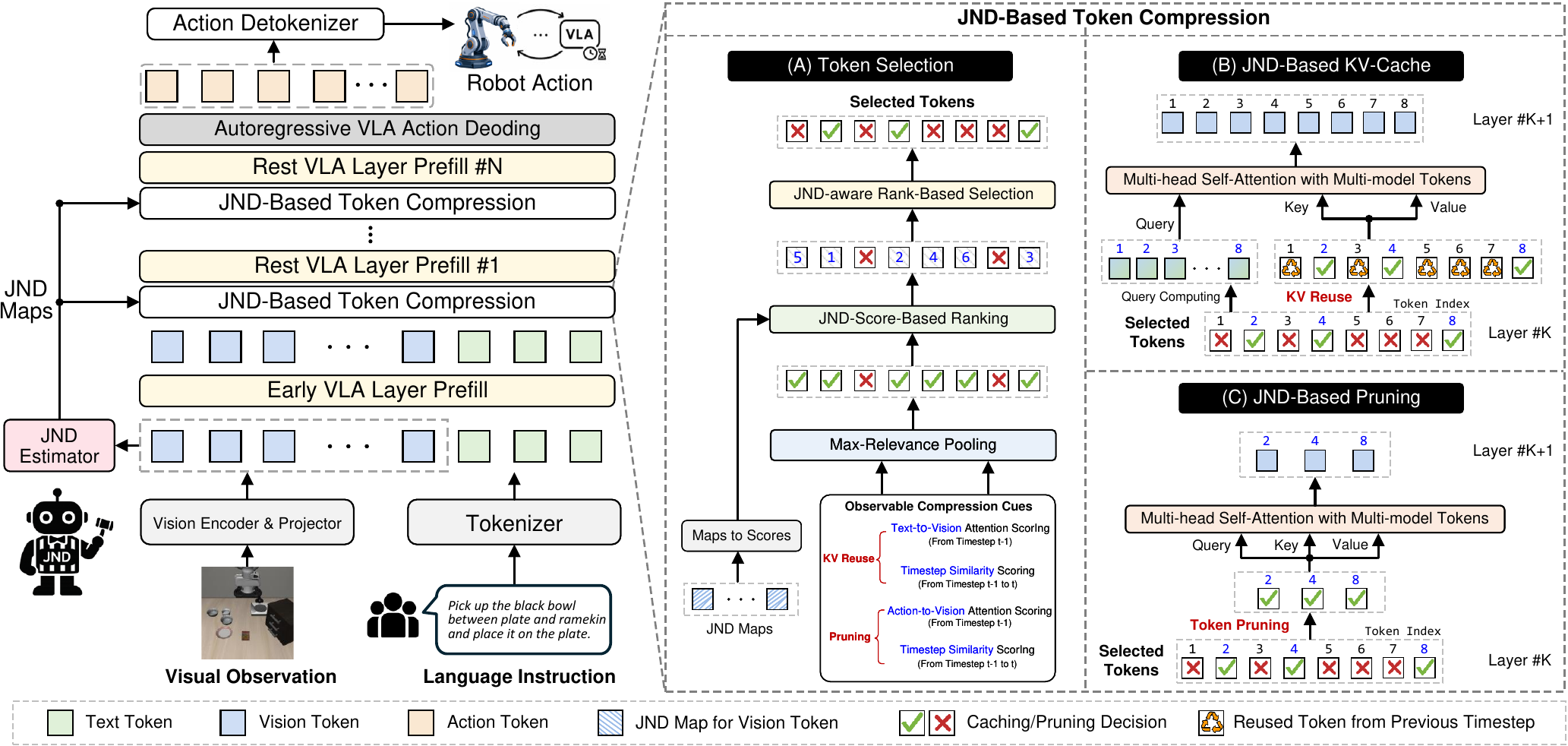}
    \vspace*{-5pt}
    \caption{\textbf{Overall framework of Action-JND-aware token compression.} Given a visual observation and a language instruction, the VLA model extracts visual tokens and language tokens for action prediction. A trainable Action-JND estimator predicts token-wise perturbation maps from the projected visual-token features. These maps are converted into action-tolerance scores that quantify the action tolerance of individual tokens. (1) The token-selection module ranks candidate visual tokens according to their JND scores, optionally together with framework-specific compression cues. (2) For KV-cache reuse, action-tolerant tokens reuse their cached key/value states from previous control steps, while action-sensitive tokens are recomputed. (3) For token pruning, highly action-tolerant tokens are removed from subsequent computation, whereas action-sensitive tokens are preserved.}
\label{fig:overall_framework}
\vspace*{-10pt}
\end{figure*}

Following the optimization strategy adopted in feature-level JND~\cite{zhao2026just}, Eq.~\eqref{eq:action_jnd_basic} can be 
relaxed into an unconstrained objective that balances perturbation magnitude and action preservation:
\begin{equation}\small
\max_{\Delta_t}\;
\lambda_{\text{mag}} M(\Delta_t)
- \lambda_{\mathrm{act}}\,
D_{\mathrm{act}}
\big(
\mathcal{R}_{\theta}(F_t,\ell),
\mathcal{R}_{\theta}(F_t+\Delta_t,\ell)
\big),
\label{eq:action_jnd_objective}
\end{equation}
where $\lambda_{\mathrm{mag}}$ and $\lambda_{\mathrm{act}}$ control the trade-off between maximizing the allowed perturbation and preserving the predicted robot action. Eq.~\eqref{eq:action_jnd_objective} provides a formulation for learning Action-JND. The objective jointly encourages large token-wise perturbations while constraining their overall effect on the policy response. The resulting perturbation magnitude therefore characterizes the token's tolerance to action-preserving perturbations and forms the basis of the learnable Action-JND estimator introduced in the next section.

\begin{figure*}[!t]
\centering
\vspace*{-15pt}
\includegraphics[width=0.85\linewidth]{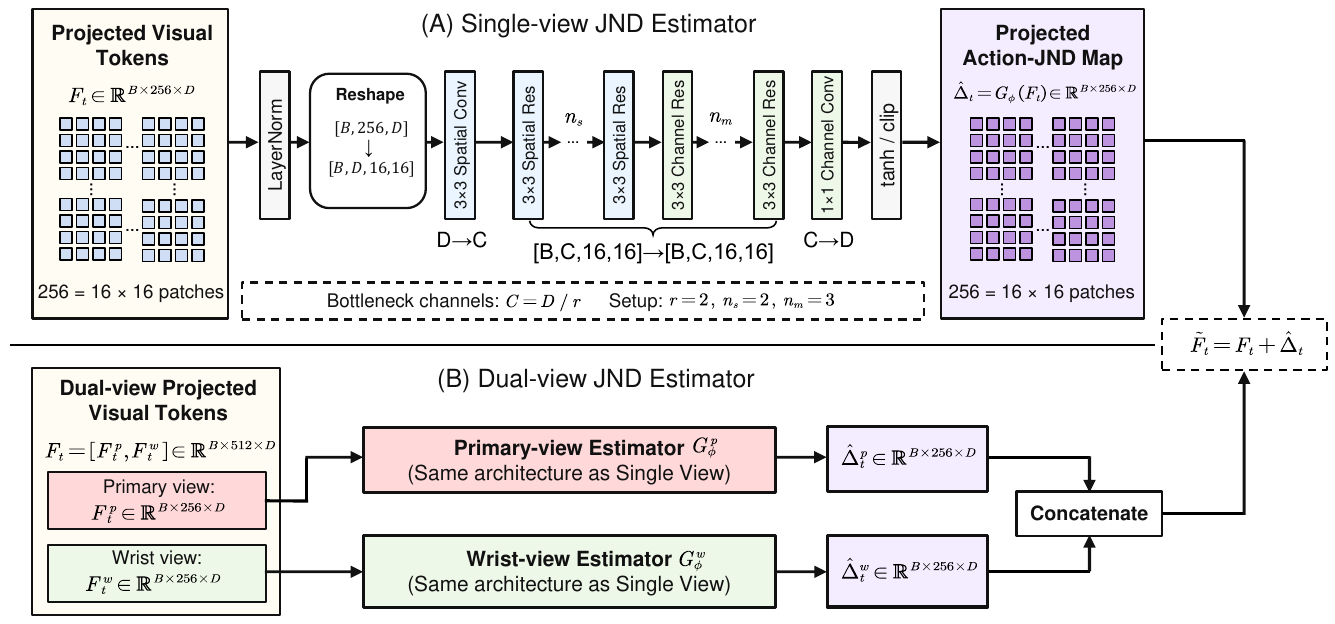}
\vspace*{-5pt}
\caption{\textbf{Architecture of the Action-JND estimator.}
For the single-view setting, projected visual tokens are reshaped into a $16 \times 16$ spatial feature map and processed by a lightweight convolutional architecture consisting of a $3 \times 3$ spatial convolution, multiple spatial residual blocks, channel-mixing residual blocks, and a $1 \times 1$ output projection. The predicted perturbations are then reshaped back into token space to form a token-wise Action-JND map. For the dual-view setting, primary-view and wrist-view tokens are processed by two independent JND estimators with the same  architecture but independent parameters, and their outputs are concatenated to produce the final Action-JND map.}
\label{fig:jnd_arch}
\vspace*{-10pt}
\end{figure*}

\section{Action-JND-Aware Token Compression}

\subsection{Overview}
After formulating Action-JND as an action-conditioned tolerance boundary, we further instantiate it as a learnable token-wise criterion for efficient VLA inference. As illustrated in Fig.~\ref{fig:overall_framework}, the proposed framework consists of three stages. First, given the projected visual-token features of a frozen VLA policy, an Action-JND estimator is trained to predict token-wise feature perturbations that are as large as possible while preserving the policy's action response. Second, during inference, the perturbation map generated by learned JND estimator is converted into token-wise action-tolerance scores. A larger score indicates that the corresponding token can tolerate a larger representation change without noticeably affecting the predicted action. Third, the learned score is used as an action-aware ranking criterion to guide token compression, including both temporal KV-cache reuse and direct visual token pruning.

The key idea is that the proposed Action-JND estimator serves as a plug-and-play
module that can be integrated into existing token compression frameworks.
It does not replace the original observable compression cues, such as temporal similarity, attention-based relevance, or token redundancy. Instead, it provides an additional criterion that directly reflects the impact of a token on the downstream robot action. Specifically, existing compression cues are first used to identify candidate
tokens, and Action-JND then ranks these candidates according to their estimated
action tolerance for subsequent KV-cache reuse or token pruning.

\subsection{Learning Token-Wise Action-JND Maps}
\label{TAP}
This subsection details how token-wise Action-JND maps are learned for VLAs. The architecture of the Action-JND estimator is first introduced, followed by the learning objective used to optimize the estimator under a frozen VLA policy.

\subsubsection{Action-JND Estimator Architecture}

Action-JND is instantiated with a lightweight estimator that predicts token-wise perturbations in the projected visual-feature space. Given visual-token features $F_t=[f_{t,1},f_{t,2},\ldots,f_{t,N}]\textcolor{blue}{^\top}  \in\mathbb{R}^{N\times D}$ extracted by the frozen VLA model, 
the estimator $G_\phi$ predicts $\hat{\Delta}_t = G_\phi(F_t)$, a learned approximation to the optimal perturbation $\Delta_t^{\star}$ in Eq.~\eqref{eq:action_jnd_basic}, with $\tilde{F}_t = F_t + \hat{\Delta}_t$.
Here, $F_t$ and $\tilde{F}_t$ denote the clean and perturbed visual-token representations, respectively. The estimator is designed to produce token-wise perturbations that are large in magnitude while remaining action-tolerable to the frozen VLA policy.

The architecture of $G_{\phi}$ is illustrated in Fig.~\ref{fig:jnd_arch}. For a single-view VLA model, the visual tokens form a $16 \times 16$ spatial grid with $N=256$. To preserve this spatial structure, $LayerNorm$ is first applied to the projected visual tokens, and the token sequence is reshaped from $[B,256,D]$ to $[B,D,16,16]$. A spatial convolution then projects the channel dimension from $D$ to $C$, where $C=D/r$ denotes the bottleneck dimension. The resulting feature map is processed by $n_s$ spatial residual blocks for local relationship modeling, followed by $n_m$ channel-mixing residual blocks for cross-channel interaction. Finally, a $1\times1$ projection maps the feature dimension back from $C$ to $D$, and the output is reshaped back to token space and bounded by a $tanh/clipping$ function.

For dual-view VLA models, the projected visual features consist of primary-view tokens and wrist-view tokens: $
F_t = [F_t^P,F_t^W]
\in \mathbb{R}^{B\times512\times D}$, where $F_t^P,F_t^W\in\mathbb{R}^{B\times256\times D}$ each correspond to a $16\times16$ spatial grid. Two JND estimators with the same architecture but independent parameters, denoted by $G_{\phi}^{P}$ and $G_{\phi}^{W}$, are used to process the two views separately:
\begin{equation}
\begin{gathered}
\hat{\Delta}_t^P = G_{\phi}^{P}(F_t^P), \,\,\,\,
\hat{\Delta}_t^W = G_{\phi}^{W}(F_t^W).
\end{gathered}
\label{eq:dual_view_estimator}
\end{equation}
The final perturbation map is obtained by concatenating the two view-specific outputs:
$
\hat{\Delta}_t
=
[\hat{\Delta}_t^P,\hat{\Delta}_t^W]
\in
\mathbb{R}^{B\times512\times D}$.
Depending on the training strategy, the estimator can also be applied only to the primary view while keeping the wrist-view tokens unperturbed. This view-specific design preserves the spatial topology of each camera view and avoids mixing different visual coordinate systems.

\begin{figure}[!t]
\centering\includegraphics[width=0.95\linewidth]{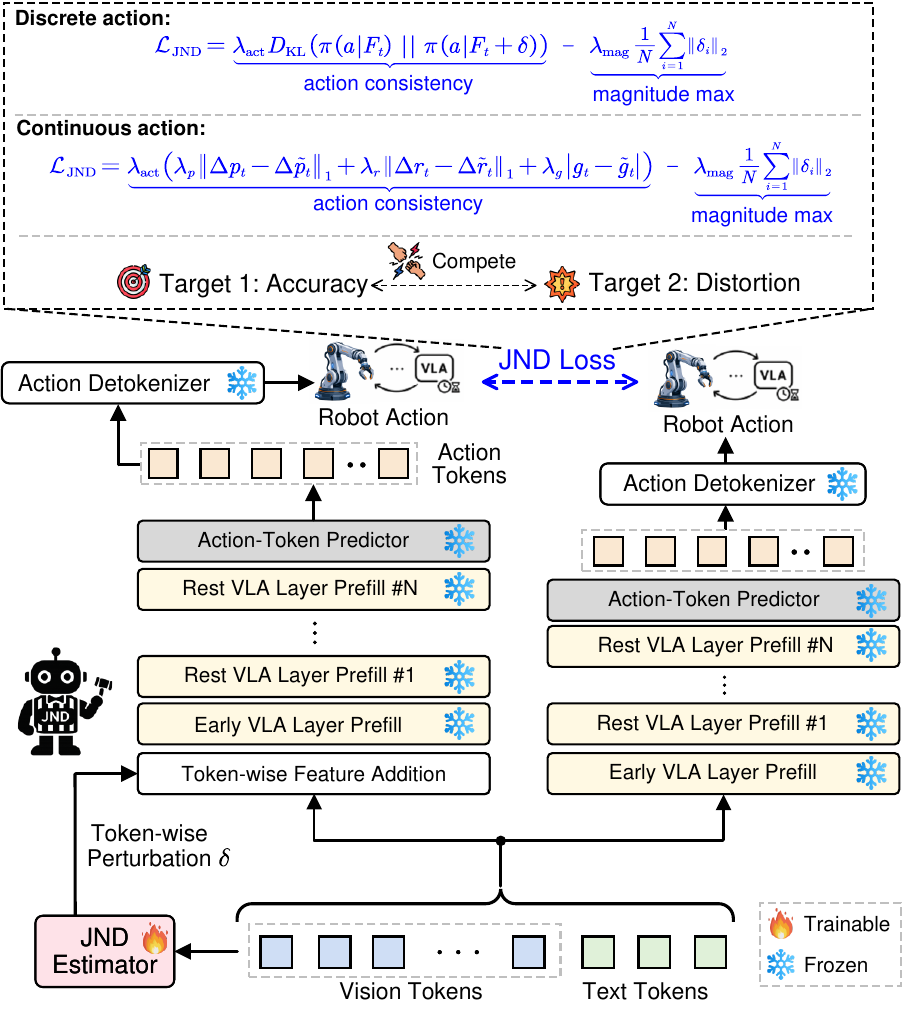}
\vspace*{-6pt}
\caption{\textbf{Training pipeline of the Action-JND estimator.} The VLA policy is frozen, and only the JND estimator is trainable. Given visual tokens and language tokens, the clean branch produces the original robot action, while the perturbed branch injects the predicted Action-JND perturbation into the visual-token features and produces a perturbed action response. The Action-JND loss encourages action consistency between the clean and perturbed branches while maximizing perturbation magnitude. The figure illustrates the KL-based objective for discrete action-token prediction; for continuous-action policies, the action discrepancy is instantiated using an action-regression distance.}
\label{fig:jnd_training}
\vspace*{-10pt}
\end{figure}

\subsubsection{Learning Objective}
Fig.~\ref{fig:jnd_training} illustrates the training procedure of the Action-JND estimator. During training, the VLA policy $\pi_{\theta}$ is kept frozen, while only the estimator $G_{\phi}$ is updated. The clean and perturbed visual features are fed into the same frozen policy to obtain two action responses: 
$r_t=\mathcal{R}_{\theta}(F_t,\ell)$,
$\tilde r_t=\mathcal{R}_{\theta}(\tilde F_t,\ell)$,
where $\ell$ denotes the language instruction. Following the unconstrained Action-JND objective in Eq.~\eqref{eq:action_jnd_objective}, the estimator is trained to encourage large perturbations while penalizing the resulting action discrepancy:
\begin{equation}
\mathcal{L}_{\mathrm{JND}}
=
\lambda_{\mathrm{act}}\cdot
D_{\mathrm{act}}
\big(
r_t,\tilde r_t
\big)
-
\lambda_{\mathrm{mag}}\cdot
M(\hat{\Delta}_t),
\label{eq:jnd_training_loss}
\end{equation}
where $\lambda_{\mathrm{act}}$ and $\lambda_{\mathrm{mag}}$ balance action consistency and perturbation magnitude. The perturbation-magnitude term is computed as
$M(\hat{\Delta}_t)
=
\frac{1}{N}
\sum_{i=1}^{N}
\|\hat{\delta}_{t,i}\|_2$,
where $\hat{\delta}_{t,i}$ denotes predicted perturbation for token $i$. This term instantiates the perturbation measure $M(\cdot)$ introduced in Eq.~\eqref{eq:receiver_jnd}.

The action discrepancy $D_{\mathrm{act}}$ is instantiated according to the action parameterization of the VLA policy. For OpenVLA-style discrete action-token prediction~\cite{kim2024openvla}, $D_{\mathrm{act}}$ is computed as the KL divergence between clean and perturbed action-token distributions. For OpenVLA-OFT-style continuous action regression~\cite{kim2025fine}, $D_{\mathrm{act}}$ is computed using an action regression distance, such as the weighted $\ell_1$ distance defined in Eq.~\eqref{eq:continuous_action_distance}. Thus, the same training framework can be applied to both discrete action-token prediction and continuous action regression, with only the action-discrepancy term changed.

\subsection{From JND Maps to Token Scores}
After training, the predicted perturbation is converted into a token-wise JND score. Two score forms can be derived:
\begin{equation}
s_{t,i}^{\mathrm{abs}}
=
\|\hat{\delta}_{t,i}\|_2,
\qquad
s_{t,i}^{\mathrm{rel}}
=
\frac{\|\hat{\delta}_{t,i}\|_2}
{\|f_{t,i}\|_2+\varepsilon},
\label{eq:jnd_score}
\end{equation}
where $\varepsilon$ is a small constant introduced for numerical stability. Both scores share the same interpretation: a larger value indicates that the corresponding token can tolerate a larger feature perturbation while preserving the policy's action response. In our implementation, the relative score $s_{t,i}^{\mathrm{rel}}$ is used for KV-cache reuse to reduce the influence of token-feature scale, whereas the absolute score $s_{t,i}^{\mathrm{abs}}$ is used for direct visual token pruning. During inference, the predicted perturbation $\hat{\Delta}_t$ is not injected into the visual features. Instead, it is used only to estimate token-wise Action-JND scores. The clean projected features $F_t$ are passed to the VLA policy for action prediction, ensuring that Action-JND affects inference only through token selection decisions.

Let $\mathcal{S}_t=\{s_{t,i}\}_{i=1}^{N}$ denote the token-wise Action-JND score map. This score serves as a ranking signal over the candidate set provided by a generic compression framework. Given an admissible candidate set $\mathcal{C}_t$, candidate tokens are ranked in descending order of their action tolerance:
\begin{equation}
\mathrm{Rank}_{\mathrm{JND}}(i)
=
\operatorname{rank}_{\downarrow}
\big(s_{t,i}\big),
\qquad i \in \mathcal{C}_t .
\label{eq:jnd_rank}
\end{equation}
Therefore, Action-JND refines the ordering among tokens that are already considered eligible for compression. When the candidate pool is insufficient, the same JND ranking can also be applied to framework-specific fallback candidates.

\begin{algorithm}[t]
\footnotesize
\caption{Action-JND-Aware KV-Cache Reuse}
\label{alg:jnd_cache}
\begin{algorithmic}[1]
\Require Current and previous camera observations $I_t$ and $I_{t-1}$;
Action-JND scores $\mathcal{S}_t=\{s_{t,i}^{\mathrm{rel}}\}_{i=1}^{N}$;
previous-step text-to-vision attention scores $\mathcal{A}_{t-1}$;
previous KV states
$\{K_{t-1}^{(l)},V_{t-1}^{(l)}\}_{l=1}^{L}$;
layer-wise reuse budgets
$\{\kappa_t^{(l)}\}_{l=1}^{L}$;
reuse policy $p$.
\Ensure Current KV states
$\{K_t^{(l)},V_t^{(l)}\}_{l=1}^{L}$.

\If{previous-step attention or KV states are unavailable}
    \State Compute all current-step KV states normally.
    \State \Return
    $\{K_t^{(l)},V_t^{(l)}\}_{l=1}^{L}$.
\EndIf

\State Estimate temporal token stability between $I_{t-1}$ and $I_t$.
\State Construct the temporally stable token set $\mathcal{T}_t$.
\State Identify task-relevant protected tokens
$\mathcal{A}_t^{\mathrm{prot}}$
using previous-step text-to-vision attention scores $\mathcal{A}_{t-1}$.

\State \colorbox{alggray}{\strut Construct safe reuse candidates:}
\Statex \hspace{\algorithmicindent}
$\mathcal{C}_{t}^{\mathrm{cache}}
=
\mathcal{T}_t
\setminus
\mathcal{A}_t^{\mathrm{prot}}$.

\For{each cache-controlled decoder layer $l=1,\ldots,L$}
    \State Determine the available safe-candidate budget:
    \Statex \hspace{\algorithmicindent}
    $\bar{\kappa}_t^{(l)}
    =
    \min
    \left(
    \kappa_t^{(l)},
    |\mathcal{C}_{t}^{\mathrm{cache}}|
    \right)$.

    \State \colorbox{alggreen}{\strut Select Action-JND-guided reuse tokens:}
    \Statex \hspace{\algorithmicindent}
    $\mathcal{R}_t^{(l)}
    =
    \operatorname{TopK}_{i\in\mathcal{C}_{t}^{\mathrm{cache}}}
    \big(
    s_{t,i}^{\mathrm{rel}},
    \bar{\kappa}_t^{(l)}
    \big)$.

    \If{$p=\texttt{strict\_guarded}$ and
    $|\mathcal{R}_t^{(l)}|<\kappa_t^{(l)}$}
        \State Construct the guarded fallback pool
        $\mathcal{B}_t$, excluding the designated
        high-attention guard positions.
        \State Fill the remaining reuse budget:
        \Statex \hspace{\algorithmicindent}
        $\mathcal{R}_t^{(l)}
        \leftarrow
        \mathcal{R}_t^{(l)}
        \cup
        \operatorname{TopK}_{i\in
        \mathcal{B}_t
        \setminus
        \mathcal{R}_t^{(l)}}
        \left(
        s_{t,i}^{\mathrm{rel}},
        \kappa_t^{(l)}-|\mathcal{R}_t^{(l)}|
        \right)$.
    \EndIf

    \For{each visual token $i\in\mathcal{R}_t^{(l)}$}
        \State Reuse stale KV states:
        \Statex \hspace{\algorithmicindent}
        $K_{t,i}^{(l)}
        \leftarrow K_{t-1,i}^{(l)}$,
        \quad
        $V_{t,i}^{(l)}
        \leftarrow V_{t-1,i}^{(l)}$.
    \EndFor

    \For{each visual token $i\notin\mathcal{R}_t^{(l)}$}
        \State Recompute $K_{t,i}^{(l)}$ and $V_{t,i}^{(l)}$
        from the current-step hidden states.
    \EndFor
\EndFor

\State \Return
$\{K_t^{(l)},V_t^{(l)}\}_{l=1}^{L}$.
\end{algorithmic}
\end{algorithm}

\subsection{JND-Aware Token Compression}
Here we present how the Action-JND scores are incorporated into token compression pipelines. The proposed criterion is not intended to redesign the compression framework. Instead, it introduces an action-aware signal into the token-selection procedure of existing methods. For a ranking-based token compression pipeline, let $\mathcal{C}_t$ denote the candidate token set generated by a generic compression framework, where the
candidate tokens may be identified based on temporal stability, attention-based
relevance, or other framework-specific cues. Given a target compression budget $\kappa_t$, the number of tokens that
can be directly selected from the candidate set is
$\bar{\kappa}_t
=
\min
\left(
\kappa_t,
|\mathcal{C}_t|
\right)$.
The Action-JND-selected token set is then computed as
$\mathcal{Q}_t
=
\operatorname{TopK}_{i\in\mathcal{C}_t}
\big(
s_{t,i},
\bar{\kappa}_t
\big)$, where tokens with larger Action-JND scores are prioritized for compression because they exhibit stronger action-level tolerance. 

This formulation provides a unified plug-and-play interface for different VLA token compression mechanisms. In this work, it is instantiated in two
representative compression scenarios: KV-cache reuse and token pruning. Although these two operations affect the VLA
inference process in different ways, they share the same underlying principle: tokens with larger JND scores can be compressed because their representations are less critical to action generation.

\textbf{JND-Aware KV-Cache Reuse.}
For KV-cache reuse, VLA-Cache~\cite{xu2026vla} is adopted as the baseline. Its goal is to reduce redundant computation across consecutive control timesteps by reusing cached key/value states for selected visual-token positions. The selected visual tokens remain in the sequence, but their key/value states are reused from the previous control timestep instead of being recomputed from the current-step visual representations. The overall  Action-JND-aware KV-cache reuse pipeline is summarized in Alg.~\ref{alg:jnd_cache}.

In the original VLA-Cache framework~\cite{xu2026vla}, token-reuse decisions are mainly guided by temporal redundancy and task relevance. Specifically, temporally stable tokens are identified by comparing adjacent visual observations, while task-relevant positions are protected according to previous-step text-to-vision attention-based filtering. These cues construct a reuse candidate set $\mathcal{C}_{t}^{\mathrm{cache}}$, which is shared across decoder layers. As shown in Alg.~\ref{alg:jnd_cache}, the Action-JND scores are applied after candidate construction to select reusable token positions. Given a layer-wise reuse budget $\kappa_t^{(l)}$, the number of tokens that can be directly selected from the candidate set is: $
\bar{\kappa}_t^{(l)}
=
\min
\left(
\kappa_t^{(l)},
\left|
\mathcal{C}_{t}^{\mathrm{cache}}
\right|
\right)$. The candidates are then ranked according to their JND scores, and the initial reuse set at decoder layer $l$ is selected as:
$
\mathcal{R}_t^{(l)}
=
\operatorname{TopK}_{i\in\mathcal{C}_{t}^{\mathrm{cache}}}
\big(
s_{t,i}^{\mathrm{rel}},
\bar{\kappa}_t^{(l)}
\big)$.
Since all decoder layers share the same ranking order, the reuse set of each layer corresponds to a different prefix of this common ordering according to the layer-wise reuse schedule.

Under the adaptive \texttt{soft} policy, only tokens from the safe candidate set are reused, and the actual reuse ratio may be lower than the requested budget when the candidate set is insufficient. For the setting with a prescribed reuse ratio, the \texttt{strict\_guarded} policy is used to enforce the target reuse budget. In this setting, when the safe candidate set cannot satisfy the desired reuse budget, the remaining quota is filled from a guarded fallback pool according to the same descending Action-JND ranking strategy, while the designated high-attention positions remain protected from fallback reuse. For each selected token $i\in\mathcal{R}_t^{(l)}$, the current key/value states are replaced by their cached states from the previous timestep:
\begin{equation}
\begin{gathered}
K_{t,i}^{(l)}
\leftarrow
K_{t-1,i}^{(l)}, \,\,\,\,
V_{t,i}^{(l)}
\leftarrow
V_{t-1,i}^{(l)},
\quad
i\in\mathcal{R}_t^{(l)} .
\end{gathered}
\label{eq:kv_reuse}
\end{equation}
Tokens outside $\mathcal{R}_t^{(l)}$ are recomputed normally from their current-step hidden states. In this way, the original compression policy and schedule of the VLA-Cache pipeline remain unchanged, while the final ranking criterion for safe and fallback candidates is replaced by an action-aware tolerance estimation. Compared with heuristic risk measures, Action-JND encourages action-tolerant tokens to reuse stale KV states while allowing action-sensitive tokens to be recomputed.

\begin{algorithm}[t]
\footnotesize
\caption{Action-JND-Aware Token Pruning}
\label{alg:jnd_pruning}
\begin{algorithmic}[1]
\Require Current and previous observations $I_t$ and $I_{t-1}$;
previous-step action-to-vision attention scores $a_{t-1}^{\mathrm{act}}$;
Action-JND scores
$\mathcal{S}_t
=
\{s_{t,i}^{\mathrm{abs}}\}_{i=1}^{N}$;
similarity threshold $\tau$;
stable-token limit $K_s$;
protection budget $K_a$;
pruning budget $\kappa_t$.
\Ensure Retained visual-token sequence $F_t^{\mathrm{keep}}$.

\State Let $\mathcal{V}=\{1,\ldots,N\}$ denote all visual-token positions.

\If{previous-step guidance is unavailable}
    \State \Return $F_t$.
\EndIf

\State Compute aligned RGB-patch similarities
$\{\rho_{t,i}\}_{i=1}^{N}$.

\State Construct the thresholded stable set:
\Statex \hspace{\algorithmicindent}
$\widetilde{\mathcal{T}}_t
=
\{i\mid\rho_{t,i}\geq\tau\}$.

\State Retain at most $K_s$ stable tokens:
\Statex \hspace{\algorithmicindent}
$\mathcal{T}_t
=
\operatorname{TopK}_{i\in\widetilde{\mathcal{T}}_t}
\left(
\rho_{t,i},
\min\left(K_s,|\widetilde{\mathcal{T}}_t|\right)
\right)$.

\State Construct the action-attention protection set:
\Statex \hspace{\algorithmicindent}
$\mathcal{A}_{t}^{\mathrm{prot}}
=
\operatorname{TopK}
\left(
a_{t-1}^{\mathrm{act}},
K_a
\right)$.

\State \colorbox{alggray}{\strut Construct safe pruning candidates:}
\Statex \hspace{\algorithmicindent}
$\mathcal{C}_{t}^{\mathrm{prune}}
=
\mathcal{T}_t
\setminus
\mathcal{A}_{t}^{\mathrm{prot}}$.

\State Select safe candidates using descending Action-JND scores:
\Statex \hspace{\algorithmicindent}
$\mathcal{P}_t
=
\operatorname{TopK}_{i\in\mathcal{C}_{t}^{\mathrm{prune}}}
\left(
s_{t,i}^{\mathrm{abs}},
\min\left(
\kappa_t,
|\mathcal{C}_{t}^{\mathrm{prune}}|
\right)
\right)$.

\If{$|\mathcal{P}_t|<\kappa_t$}
    \State \colorbox{alggreen}{\strut Fill the remaining pruning budget:}
    \Statex \hspace{\algorithmicindent}
    $\mathcal{P}_t
    \leftarrow
    \mathcal{P}_t
    \cup
    \operatorname{TopK}_{i\in\mathcal{V}\setminus\mathcal{P}_t}
    \left(
    s_{t,i}^{\mathrm{abs}},
    \kappa_t-|\mathcal{P}_t|
    \right)$.
\EndIf

\State Physically remove the selected visual tokens:
\Statex \hspace{\algorithmicindent}
$F_t^{\mathrm{keep}}
=
\left[
f_{t,i}
\mid
i\notin\mathcal{P}_t
\right]$.

\State Feed $F_t^{\mathrm{keep}}$ into decoder layer index 3
and subsequent layers.

\State \Return $F_t^{\mathrm{keep}}$.
\end{algorithmic}
\end{algorithm}

\textbf{JND-Aware Token Pruning.}
For token pruning, we adapt the safe-candidate construction strategy of
VLA-Cache~\cite{xu2026vla} from stale-KV reuse to direct token
removal. Visual tokens that are temporally stable and receive low
attention during previous action generation are considered potential 
pruning candidates. Under a fixed pruning budget, the proposed method
selects candidates according to their learned Action-JND scores and
removes the most action-tolerant ones from the multimodal sequence before subsequent decoder layers. Alg.~\ref{alg:jnd_pruning} summarizes the resulting procedure under the \texttt{budgeted} policy.

At control step $t$, let $p_{t,i}$ and $p_{t-1,i}$ denote the aligned RGB patches corresponding to visual token $i$ in the current and previous observations, respectively. Their temporal similarity is computed as
$
\rho_{t,i}
=
\operatorname{cos}
\big(
p_{t,i},p_{t-1,i}
\big)$.
To avoid generating an excessively large candidate pool, tokens whose similarity exceeds a threshold $\tau$ form an initial stable set. When the number of stable tokens exceeds the prescribed upper bound $K_s$, only $K_s$ tokens with the highest similarities are retained:
\begin{equation}\footnotesize
\begin{aligned}
\widetilde{\mathcal{T}}_t
=
\left\{
i\mid\rho_{t,i}\geq\tau
\right\}, \quad
\mathcal{T}_t
=
\operatorname{TopK}_{i\in\widetilde{\mathcal{T}}_t}
\left(
\rho_{t,i},
\min\left(K_s,|\widetilde{\mathcal{T}}_t|\right)
\right).
\end{aligned}
\label{eq:stable_pruning_tokens}
\end{equation}

To protect tokens associated with action generation process, the $K_a$ positions receiving the highest action-to-vision attention at the previous control timestep are collected as
$
\mathcal{A}_{t}^{\mathrm{prot}}
=
\operatorname{TopK}
\big(
a_{t-1}^{\mathrm{act}},K_a
\big)$.
Since token pruning is performed at an intermediate layer
during prefilling, we record the previous-timestep action-to-vision
attention from layers preceding the pruning layer during action decoding,
where the complete $N$-token visual grid remains available. The safe pruning candidate pool is then constructed as $\mathcal{C}_{t}^{\mathrm{prune}}
=
\mathcal{T}_t
\setminus
\mathcal{A}_{t}^{\mathrm{prot}}$, where $\setminus$ means token removal.
Let $K_t$ denote the target number of retained tokens, and
$\kappa_t=N-K_t$ the corresponding number of tokens to be removed.
The candidate tokens are ranked in descending order of $s_{t,i}^{\mathrm{abs}}$, such that tokens with greater action tolerance
are removed first. Under the \texttt{budgeted} policy, which enforces a fixed token budget, the final pruning set is,
\begin{equation}\footnotesize
\mathcal{P}_t
=
\begin{cases}
\displaystyle
\operatorname{TopK}_{i\in\mathcal{C}_{t}^{\mathrm{prune}}}
\big(
s_{t,i}^{\mathrm{abs}},
\kappa_t
\big),
&
|\mathcal{C}_{t}^{\mathrm{prune}}|
\geq\kappa_t,
\\[3mm]
\displaystyle
\mathcal{C}_{t}^{\mathrm{prune}}
\cup
\operatorname{TopK}_{i\notin\mathcal{C}_{t}^{\mathrm{prune}}}
\big(
s_{t,i}^{\mathrm{abs}},
\kappa_t-|\mathcal{C}_{t}^{\mathrm{prune}}|
\big),
&
|\mathcal{C}_{t}^{\mathrm{prune}}|
<\kappa_t.
\end{cases}
\label{eq:jnd_budgeted_pruning}
\end{equation}
This formulation ensures that the safe candidate pool is always prioritized. When the candidate pool contains fewer than $\kappa_t$ tokens, all safe candidates are pruned first, and the remaining pruning quota is filled from the other visual-token
positions according to the same descending Action-JND ranking. Although these additional pruning candidates may fall outside the original protection constraints, selecting them according to Action-JND still prioritizes tokens with relatively high action tolerance, thereby reducing the risk of affecting the downstream robot action. 

\begin{table*}[t]
\centering
\small
\vspace{-1em}
\setlength{\tabcolsep}{6.0pt}
\renewcommand{\arraystretch}{1.0}
\caption{Performance of Action-JND and KV-cache baselines on OpenVLA
on the LIBERO benchmark~\cite{liu2023libero}. Vanilla OpenVLA serves
as the uncompressed reference. For each reuse setting, success rates
(\%) on \emph{Spatial}, \emph{Object}, \emph{Goal}, and \emph{Long}
and their mean (\emph{Acc.}) are reported. $\Delta$Acc. denotes the
accuracy difference relative to VLA-Cache~\cite{xu2026vla} under the same
setting. FLOPs, relative FLOPs, \\ CUDA latency, latency difference relative to
vanilla OpenVLA, and control frequency are also reported. \textcolor{blue}{Blue} and \textcolor{red}{red} \\ indicate the best and worst results, respectively, within
each reuse setting according to\\ the indicated $\uparrow/\downarrow$
direction, while pink rows highlight Action-JND}
\vspace{-0.6em}
\label{tab:openvla-action-jnd-baselines}
\resizebox{\textwidth}{!}{%
\begin{tabular}{l|cccc|ccccccc}
\toprule[1.2pt]
\multirow{2}{*}{Method}
& \multicolumn{11}{c}{OpenVLA} \\
\cmidrule(lr){2-12}
& Spatial & Object & Goal & Long
& Acc.(\%)$\uparrow$
& $\Delta$Acc.$\uparrow$
& FLOPs(T)$\downarrow$
& Rel. FLOPs(\%)$\downarrow$
& Latency(ms)$\downarrow$
& $\Delta$Latency(ms)$\downarrow$
& Freq.(Hz)$\uparrow$ \\
\midrule

\rowcolor{gray!15}
\multicolumn{12}{c}{\emph{Upper Bound (Vanilla, 100\%)}} \\
Vanilla
& 81.40 & 64.40 & 79.20 & 57.20 & \textbf{70.55}
& -- & 1.864 & 100.00\% & 46.07 & -- & 21.71 \\

\midrule
\rowcolor{gray!15}
\multicolumn{12}{c}{\emph{Soft Adaptive KV Cache}} \\
FastV~\cite{chen2024image}

& \textcolor{posblue}{79.80} & 60.20 & 74.60 & 48.00 & \textbf{65.65}
& \textbf{+1.55} & \textcolor{negred}{1.446} & 77.58\% & 40.95 & -5.12 & 24.42 \\
SparseVLM~\cite{zhang2024sparsevlm}
& 77.80 & 60.20 & \textcolor{posblue}{76.40} & 48.80 & \textbf{65.80}
& \textbf{+1.70} & 1.446 & \textcolor{negred}{77.59\%} & 41.57 & -4.50 & 24.06 \\
DivPrune~\cite{alvar2025divprune}
& 78.60 & 61.80 & 75.20 & 48.20 & \textbf{65.95}
& \textbf{+1.85} & 1.434 & 76.93\% & \textcolor{negred}{43.20} & \textcolor{negred}{-2.88} & \textcolor{negred}{23.19} \\
VLA-Cache~\cite{xu2026vla}
& \textcolor{negred}{77.60} & \textcolor{negred}{58.20} & 75.40 & \textcolor{negred}{45.20} & \textcolor{negred}{\textbf{64.10}}
& \textcolor{negred}{\textbf{+0.00}} & \textcolor{posblue}{1.431} & \textcolor{posblue}{76.75\%} & 41.68 & -4.39 & 24.02 \\
\rowcolor{actpink}
\textbf{Action-JND (Ours)} & 79.60 & \textcolor{posblue}{64.40} & \textcolor{negred}{74.40} & \textcolor{posblue}{50.00} & \textcolor{posblue}{\textbf{67.10}} & \textcolor{posblue}{\textbf{+3.00}} & 1.445 & 77.52\% & \textcolor{posblue}{40.71} & \textcolor{posblue}{-5.36} & \textcolor{posblue}{24.57} \\

\midrule
\rowcolor{gray!15}
\multicolumn{12}{c}{\emph{KV Cache Reuse Ratio = 30\%}} \\
FastV~\cite{chen2024image}
& 77.60 & 60.40 & 74.00 & 46.00 & \textcolor{negred}{\textbf{64.50}}
& \textcolor{negred}{\textbf{-0.20}} & \textcolor{posblue}{1.339} & \textcolor{posblue}{71.83\%} & 43.79 & -2.29 & 22.86 \\
SparseVLM~\cite{zhang2024sparsevlm}
& 77.60 & 61.40 & \textcolor{negred}{73.40} & 47.00 & \textbf{64.85}
& \textbf{+0.15} & 1.339 & 71.84\% & \textcolor{negred}{47.15} & \textcolor{negred}{+1.07} & \textcolor{negred}{21.25} \\
DivPrune~\cite{alvar2025divprune}
& 77.60 & 61.60 & \textcolor{posblue}{76.80} & 45.60 & \textbf{65.40}
& \textbf{+0.70} & 1.339 & 71.85\% & 39.73 & -6.34 & 25.19 \\
VLA-Cache~\cite{xu2026vla}
& \textcolor{negred}{77.40} & \textcolor{negred}{60.20} & 76.00 & \textcolor{negred}{45.20} & \textbf{64.70}
& \textbf{+0.00} & 1.339 & 71.84\% & 39.85 & -6.23 & 25.11 \\
\rowcolor{actpink}
\textbf{Action-JND (Ours)} & \textcolor{posblue}{80.80} & \textcolor{posblue}{62.20} & 75.60 & \textcolor{posblue}{49.80} & \textcolor{posblue}{\textbf{67.10}} & \textcolor{posblue}{\textbf{+2.40}} & \textcolor{negred}{1.340} & \textcolor{negred}{71.90\%} & \textcolor{posblue}{38.55} & \textcolor{posblue}{-7.52} & \textcolor{posblue}{25.95} \\

\midrule
\rowcolor{gray!15}
\multicolumn{12}{c}{\emph{KV Cache Reuse Ratio = 40\%}} \\
FastV~\cite{chen2024image}
& \textcolor{negred}{76.00} & 60.60 & 73.60 & 46.40 & \textbf{64.15}
& \textbf{-1.05} & \textcolor{posblue}{1.158} & \textcolor{posblue}{62.10\%} & \textcolor{negred}{42.86} & \textcolor{negred}{-3.21} & \textcolor{negred}{23.34} \\
SparseVLM~\cite{zhang2024sparsevlm}
& 76.00 & \textcolor{negred}{58.40} & \textcolor{negred}{73.40} & \textcolor{negred}{43.40} & \textcolor{negred}{\textbf{62.80}}
& \textcolor{negred}{\textbf{-2.40}} & 1.158 & 62.10\% & 37.33 & -8.74 & 26.79 \\
DivPrune~\cite{alvar2025divprune}
& 77.00 & 60.20 & 76.80 & 46.60 & \textbf{65.15}
& \textbf{-0.05} & \textcolor{negred}{1.162} & 62.32\% & 36.66 & -9.41 & 27.28 \\
VLA-Cache~\cite{xu2026vla}
& \textcolor{posblue}{78.60} & \textcolor{posblue}{62.40} & 74.20 & 45.60 & \textbf{65.20}
& \textbf{+0.00} & 1.160 & 62.24\% & 37.85 & -8.22 & 26.42 \\
\rowcolor{actpink}
\textbf{Action-JND (Ours)} & 77.80 & 62.20 & \textcolor{posblue}{78.60} & \textcolor{posblue}{47.60} & \textcolor{posblue}{\textbf{66.55}} & \textcolor{posblue}{\textbf{+1.35}} & 1.162 & \textcolor{negred}{62.33\%} & \textcolor{posblue}{36.20} & \textcolor{posblue}{-9.87} & \textcolor{posblue}{27.62} \\

\midrule
\rowcolor{gray!15}
\multicolumn{12}{c}{\emph{KV Cache Reuse Ratio = 60\%}} \\
FastV~\cite{chen2024image}
& 41.20 & 41.60 & 64.20 & 10.20 & \textbf{39.30}
& \textbf{+6.10} & \textcolor{posblue}{0.785} & \textcolor{posblue}{42.13\%} & 31.54 & -14.53 & 31.71 \\
SparseVLM~\cite{zhang2024sparsevlm}
& 38.60 & 37.60 & 64.60 & 12.80 & \textbf{38.40}
& \textbf{+5.20} & 0.785 & 42.13\% & 31.87 & -14.20 & 31.39 \\
DivPrune~\cite{alvar2025divprune}
& 40.20 & 49.00 & 70.60 & 17.00 & \textbf{44.20}
& \textbf{+11.00} & 0.791 & 42.44\% & 31.88 & -14.19 & 31.38 \\
VLA-Cache~\cite{xu2026vla}
& \textcolor{negred}{31.40} & \textcolor{negred}{30.60} & \textcolor{negred}{63.20} & \textcolor{negred}{7.60} & \textcolor{negred}{\textbf{33.20}}
& \textcolor{negred}{\textbf{+0.00}} & 0.792 & 42.49\% & \textcolor{negred}{32.36} & \textcolor{negred}{-13.71} & \textcolor{negred}{30.91} \\
\rowcolor{actpink}
\textbf{Action-JND (Ours)} & \textcolor{posblue}{71.20} & \textcolor{posblue}{60.80} & \textcolor{posblue}{71.00} & \textcolor{posblue}{24.60} & \textcolor{posblue}{\textbf{56.90}} & \textcolor{posblue}{\textbf{+23.70}} & \textcolor{negred}{0.793} & \textcolor{negred}{42.54\%} & \textcolor{posblue}{30.18} & \textcolor{posblue}{-15.89} & \textcolor{posblue}{33.14} \\

\midrule
\rowcolor{gray!15}
\multicolumn{12}{c}{\emph{KV Cache Reuse Ratio = 80\%}} \\
FastV~\cite{chen2024image}
& 32.40 & 33.80 & 61.00 & 8.00 & \textbf{33.80}
& \textbf{+20.00} & \textcolor{posblue}{0.742} & \textcolor{posblue}{39.83\%} & 43.34 & -2.73 & 23.19 \\
SparseVLM~\cite{zhang2024sparsevlm}
& 33.60 & 35.80 & 61.80 & 10.40 & \textbf{35.40}
& \textbf{+21.60} & 0.743 & 39.84\% & \textcolor{negred}{43.69} & \textcolor{negred}{-2.38} & \textcolor{negred}{22.91} \\
DivPrune~\cite{alvar2025divprune}
& 31.00 & 40.80 & 63.20 & 14.60 & \textbf{37.40}
& \textbf{+23.60} & 0.745 & 39.96\% & \textcolor{posblue}{31.13} & \textcolor{posblue}{-14.94} & \textcolor{posblue}{32.13} \\
VLA-Cache~\cite{xu2026vla}
& \textcolor{negred}{10.00} & \textcolor{negred}{7.60} & \textcolor{negred}{35.80} & \textcolor{negred}{1.80} & \textcolor{negred}{\textbf{13.80}}
& \textcolor{negred}{\textbf{+0.00}} & 0.749 & 40.21\% & 31.69 & -14.38 & 31.56 \\
\rowcolor{actpink}
\textbf{Action-JND (Ours)} & \textcolor{posblue}{69.40} & \textcolor{posblue}{58.80} & \textcolor{posblue}{69.20} & \textcolor{posblue}{24.40} & \textcolor{posblue}{\textbf{55.45}} & \textcolor{posblue}{\textbf{+41.65}} & \textcolor{negred}{0.754} & \textcolor{negred}{40.44\%} & 34.79 & -11.28 & 28.81 \\

\bottomrule[1.2pt]
\end{tabular}%
}
\vspace{-0.7em}
\end{table*}

\section{Experiments}
\subsection{Experimental Setup}

\textbf{Benchmarks and backbones.}
Experiments are conducted on the LIBERO benchmark~\cite{liu2023libero}, a commonly used benchmark for evaluating language-conditioned robotic manipulation and VLA policy learning. Following standard VLA evaluation protocols~\cite{kim2025fine,xu2026vla,liu2025vla}, four standard LIBERO suites are used for evaluation: \emph{Spatial}, \emph{Object}, \emph{Goal}, and \emph{Long}, which cover spatial reasoning, object-centric manipulation, goal-conditioned execution, and long-horizon tasks, respectively. The proposed Action-JND is evaluated on two representative VLA backbones: OpenVLA~\cite{kim2024openvla} and OpenVLA-OFT~\cite{kim2025fine}. OpenVLA uses a single RGB view and predicts discretized action tokens. OpenVLA-OFT predicts continuous action chunks and can take both primary-view and wrist-view observations.

\textbf{Training configuration.}
For OpenVLA, the JND estimator is optimized with the action-token KL objective in Eq.~\eqref{eq:discrete_action_distance}, using
$\lambda_{\mathrm{act}}=1.0$ and
$\lambda_{\mathrm{mag}}\in\{0.02,0.01,0.005\}$. For OpenVLA-OFT, the primary- and wrist-view estimators are trained separately with the continuous action-regression objective in
Eq.~\eqref{eq:continuous_action_distance}, perturbing one view while keeping the other clean. We set $\lambda_{\mathrm{act}}=1.0$,
$\lambda_{\mathrm{mag}}=3\times10^{-4}$ for  primary view, while
sweeping $\lambda_{\mathrm{mag}}\in\{3,5,7,8\}\times10^{-4}$ for 
wrist view. 

\begin{table*}[t]
\centering
\small
\vspace{-2em}
\setlength{\tabcolsep}{6.0pt}
\renewcommand{\arraystretch}{1.0}
\caption{Performance of Action-JND and KV-cache baselines on
OpenVLA-OFT on the LIBERO benchmark~\cite{liu2023libero}.
Vanilla OpenVLA-OFT serves as the uncompressed reference. For each
reuse setting, suite-wise and average success rates are reported, with
$\Delta$Acc. denoting the accuracy difference from
VLA-Cache~\cite{xu2026vla} under the same setting. FLOPs, relative
FLOPs, CUDA latency,\\ latency difference from vanilla OpenVLA-OFT, and
control frequency are also reported. The color annotations follow Table~\ref{tab:openvla-action-jnd-baselines}}
\vspace{-0.5em}
\label{tab:openvla-oft-action-jnd-baselines}
\resizebox{\textwidth}{!}{%
\begin{tabular}{l|cccc|ccccccc}
\toprule[1.2pt]
\multirow{2}{*}{Method}
& \multicolumn{11}{c}{OpenVLA-OFT} \\
\cmidrule(lr){2-12}
& Spatial & Object & Goal & Long
& Acc.(\%)$\uparrow$
& $\Delta$Acc.$\uparrow$
& FLOPs(T)$\downarrow$
& Rel. FLOPs(\%)$\downarrow$
& Latency(ms)$\downarrow$
& $\Delta$Latency(ms)$\downarrow$
& Freq.(Hz)$\uparrow$ \\
\midrule

\rowcolor{gray!15}
\multicolumn{12}{c}{\emph{Upper Bound (Vanilla, 100\%)}} \\
Vanilla
& 94.20 & 97.60 & 95.80 & 92.60 & \textbf{95.05}
& -- & 3.970 & 100.00\% & 137.43 & -- & 58.23 \\

\midrule
\rowcolor{gray!15}
\multicolumn{12}{c}{\emph{Soft Adaptive KV Cache}} \\
FastV~\cite{chen2024image}
& \textcolor{negred}{94.20} & 97.60 & 96.80 & 93.20 & \textbf{95.45}
& \textbf{+0.10} & 3.244 & 81.72\% & 132.30 & -5.13 & 60.51 \\
SparseVLM~\cite{zhang2024sparsevlm}
& 94.40 & \textcolor{negred}{97.40} & 96.40 & \textcolor{negred}{93.00} & \textcolor{negred}{\textbf{95.30}}
& \textcolor{negred}{\textbf{-0.05}} & 3.245 & 81.73\% & \textcolor{negred}{139.58} & \textcolor{negred}{+2.15} & \textcolor{negred}{57.40} \\
DivPrune~\cite{alvar2025divprune}
& \textcolor{posblue}{95.00} & 97.40 & 96.00 & 93.80 & \textbf{95.55}
& \textbf{+0.20} & \textcolor{negred}{3.251} & \textcolor{negred}{81.88\%} & 135.04 & -2.39 & 59.27 \\
VLA-Cache~\cite{xu2026vla}
& 94.40 & \textcolor{posblue}{97.80} & \textcolor{negred}{95.60} & 93.60 & \textbf{95.35}
& \textbf{+0.00} & \textcolor{posblue}{3.239} & \textcolor{posblue}{81.59\%} & \textcolor{posblue}{106.44} & \textcolor{posblue}{-30.99} & \textcolor{posblue}{75.19} \\
\rowcolor{actpink}
\textbf{Action-JND (Ours)} & 94.60 & 97.60 & \textcolor{posblue}{97.60} & \textcolor{posblue}{95.00} & \textcolor{posblue}{\textbf{96.20}} & \textcolor{posblue}{\textbf{+0.85}} & 3.242 & 81.67\% & 111.63 & -25.80 & 72.09 \\

\midrule
\rowcolor{gray!15}
\multicolumn{12}{c}{\emph{KV Cache Reuse Ratio = 30\%}} \\
FastV~\cite{chen2024image}
& \textcolor{posblue}{94.60} & 97.60 & \textcolor{negred}{95.60} & 93.00 & \textcolor{negred}{\textbf{95.20}}
& \textcolor{negred}{\textbf{-0.25}} & \textcolor{posblue}{2.991} & 75.35\% & 126.46 & -10.97 & 63.27 \\
SparseVLM~\cite{zhang2024sparsevlm}
& \textcolor{negred}{93.60} & \textcolor{negred}{97.20} & 97.00 & \textcolor{posblue}{94.40} & \textcolor{posblue}{\textbf{95.55}}
& \textcolor{posblue}{\textbf{+0.10}} & 2.991 & 75.35\% & 127.29 & -10.13 & 62.86 \\
DivPrune~\cite{alvar2025divprune}
& 94.40 & \textcolor{posblue}{98.40} & \textcolor{posblue}{97.40} & \textcolor{negred}{91.80} & \textbf{95.50}
& \textbf{+0.05} & \textcolor{negred}{2.992} & \textcolor{negred}{75.37\%} & \textcolor{negred}{130.14} & \textcolor{negred}{-7.29} & \textcolor{negred}{61.47} \\
VLA-Cache~\cite{xu2026vla}
& 93.60 & 98.40 & 96.80 & 93.00 & \textbf{95.45}
& \textbf{+0.00} & 2.991 & \textcolor{posblue}{75.34\%} & 121.26 & -16.17 & 65.99 \\
\rowcolor{actpink}
\textbf{Action-JND (Ours)} & 94.20 & 98.00 & 96.80 & 93.00 & \textbf{95.50} & \textbf{+0.05} & 2.992 & 75.37\% & \textcolor{posblue}{106.83} & \textcolor{posblue}{-30.60} & \textcolor{posblue}{75.46} \\

\midrule
\rowcolor{gray!15}
\multicolumn{12}{c}{\emph{KV Cache Reuse Ratio = 40\%}} \\
FastV~\cite{chen2024image}
& \textcolor{posblue}{94.60} & 97.40 & 96.80 & \textcolor{negred}{92.20} & \textbf{95.25}
& \textbf{-0.50} & \textcolor{posblue}{2.661} & 67.04\% & 122.22 & -15.21 & 65.51 \\
SparseVLM~\cite{zhang2024sparsevlm}
& 94.20 & \textcolor{negred}{97.00} & \textcolor{negred}{96.40} & 92.60 & \textcolor{negred}{\textbf{95.05}}
& \textcolor{negred}{\textbf{-0.70}} & 2.661 & 67.04\% & \textcolor{negred}{131.86} & \textcolor{negred}{-5.57} & \textcolor{negred}{60.67} \\
DivPrune~\cite{alvar2025divprune}
& \textcolor{negred}{93.80} & 97.60 & 96.80 & 93.40 & \textbf{95.40}
& \textbf{-0.35} & 2.662 & 67.06\% & 118.96 & -18.46 & 67.31 \\
VLA-Cache~\cite{xu2026vla}
& 94.00 & \textcolor{posblue}{98.20} & \textcolor{posblue}{97.20} & \textcolor{posblue}{93.60} & \textcolor{posblue}{\textbf{95.75}}
& \textcolor{posblue}{\textbf{+0.00}} & 2.661 & \textcolor{posblue}{67.03\%} & 112.75 & -24.68 & 70.96 \\
\rowcolor{actpink}
\textbf{Action-JND (Ours)} & 94.20 & 97.60 & 96.40 & 92.40 & \textbf{95.15} & \textbf{-0.60} & \textcolor{negred}{2.664} & \textcolor{negred}{67.09\%} & \textcolor{posblue}{100.91} & \textcolor{posblue}{-36.52} & \textcolor{posblue}{80.37} \\

\midrule
\rowcolor{gray!15}
\multicolumn{12}{c}{\emph{KV Cache Reuse Ratio = 60\%}} \\
FastV~\cite{chen2024image}
& 89.40 & \textcolor{negred}{80.80} & 97.20 & 77.60 & \textbf{86.25}
& \textbf{-0.25} & 2.001 & 50.40\% & 108.14 & -29.29 & 75.51 \\
SparseVLM~\cite{zhang2024sparsevlm}
& 90.00 & 81.20 & 96.00 & 77.20 & \textcolor{negred}{\textbf{86.10}}
& \textcolor{negred}{\textbf{-0.40}} & 2.001 & 50.41\% & \textcolor{negred}{109.68} & \textcolor{negred}{-27.75} & \textcolor{negred}{73.12} \\
DivPrune~\cite{alvar2025divprune}
& 90.60 & 92.80 & 96.00 & \textcolor{posblue}{85.20} & \textbf{91.15}
& \textbf{+4.65} & 2.006 & 50.53\% & 106.81 & -30.62 & 75.16 \\
VLA-Cache~\cite{xu2026vla}
& \textcolor{negred}{89.00} & 88.20 & \textcolor{negred}{94.40} & \textcolor{negred}{74.40} & \textbf{86.50}
& \textbf{+0.00} & \textcolor{posblue}{1.993} & \textcolor{posblue}{50.20\%} & 100.38 & -37.05 & 79.93 \\
\rowcolor{actpink}
\textbf{Action-JND (Ours)} & \textcolor{posblue}{92.00} & \textcolor{posblue}{96.80} & \textcolor{posblue}{97.40} & 84.60 & \textcolor{posblue}{\textbf{92.70}} & \textcolor{posblue}{\textbf{+6.20}} & \textcolor{negred}{2.010} & \textcolor{negred}{50.63\%} & \textcolor{posblue}{90.45} & \textcolor{posblue}{-46.98} & \textcolor{posblue}{90.53} \\

\midrule
\rowcolor{gray!15}
\multicolumn{12}{c}{\emph{KV Cache Reuse Ratio = 80\%}} \\
FastV~\cite{chen2024image}
& 89.20 & 76.00 & 95.40 & 76.00 & \textbf{84.15}
& \textbf{+4.70} & 1.974 & 49.72\% & 108.17 & -29.26 & 75.61 \\
SparseVLM~\cite{zhang2024sparsevlm}
& 90.20 & 76.20 & 95.60 & 78.00 & \textbf{85.00}
& \textbf{+5.55} & 1.974 & 49.73\% & \textcolor{negred}{113.61} & \textcolor{negred}{-23.82} & \textcolor{negred}{70.96} \\
DivPrune~\cite{alvar2025divprune}
& 90.80 & \textcolor{posblue}{81.40} & 95.60 & 83.40 & \textbf{87.80}
& \textbf{+8.35} & \textcolor{negred}{1.981} & \textcolor{negred}{49.91\%} & 107.54 & -29.89 & 74.72 \\
VLA-Cache~\cite{xu2026vla}
& \textcolor{negred}{88.80} & \textcolor{negred}{66.20} & \textcolor{negred}{92.80} & \textcolor{negred}{70.00} & \textcolor{negred}{\textbf{79.45}}
& \textcolor{negred}{\textbf{+0.00}} & \textcolor{posblue}{1.962} & \textcolor{posblue}{49.42\%} & 100.10 & -37.33 & 79.94 \\
\rowcolor{actpink}
\textbf{Action-JND (Ours)} & \textcolor{posblue}{92.00} & 78.20 & \textcolor{posblue}{97.60} & \textcolor{posblue}{90.40} & \textcolor{posblue}{\textbf{89.55}} & \textcolor{posblue}{\textbf{+10.10}} & 1.963 & 49.45\% & \textcolor{posblue}{91.05} & \textcolor{posblue}{-46.38} & \textcolor{posblue}{89.45} \\

\bottomrule[1.2pt]
\end{tabular}%
}\vspace{-0.7em}
\end{table*}

\textbf{Baselines.}
For KV-cache reuse, we compare Action-JND with VLA-Cache~\cite{xu2026vla} and KV-cache-compatible adaptations of
FastV~\cite{chen2024image}, SparseVLM~\cite{zhang2024sparsevlm},
and DivPrune~\cite{alvar2025divprune}. For token pruning, we compare Action-JND with FastV~\cite{chen2024image}, SparseVLM~\cite{zhang2024sparsevlm},
DivPrune~\cite{alvar2025divprune}, and a VLA-Cache-based pruning
baseline~\cite{xu2026vla}, which transfers the candidate construction
and proxy-risk ranking of VLA-Cache to  visual-token pruning.

\textbf{Compression settings.}
For KV-cache reuse, we evaluate the adaptive \texttt{soft} policy, whose actual reuse ratio depends on the available safe candidates, and the \texttt{strict\_guarded} policy with target reuse ratios of $30\%$, $40\%$, $60\%$, and $80\%$. For token pruning, we evaluate the \texttt{budgeted} policy with target pruning ratios of $25\%$, $50\%$, $75\%$, and $87.5\%$. 

\textbf{Implementation environments.}
The KV-cache experiments are built on the publicly released VLA-Cache~\cite{xu2026vla} codebase, whereas the token-pruning-related experiments are
built upon the official OpenVLA codebase~\cite{kim2024openvla}. Since the two codebases rely on
different software dependencies, the reported vanilla OpenVLA results of different compression tasks
may differ.

\textbf{Evaluation metrics.}
Task performance is evaluated using the success rates on each 
LIBERO suite and their mean, denoted as average accuracy
(\emph{Acc.}). $\Delta$Acc. measures the accuracy difference
relative to the corresponding baseline under the same compression
setting. For efficiency, FLOPs quantify the theoretical computational
cost, while relative FLOPs report the remaining computation as a
percentage of the vanilla model. CUDA latency measures the actual GPU execution time for action generation,
$\Delta$Latency denotes its difference relative to the vanilla model, and
control frequency measures how frequently the policy can update robot actions during closed-loop control. A higher control frequency enables more frequent action updates during
closed-loop execution.

\begin{table*}[t]
\centering
\small
\setlength{\tabcolsep}{6.0pt}
\renewcommand{\arraystretch}{1.0}
\vspace{-2em}
\caption{Performance of Action-JND and token-pruning baselines on
OpenVLA on the LIBERO benchmark~\cite{liu2023libero}. Vanilla OpenVLA
serves as the uncompressed reference. For each pruning setting,
suite-wise and average success rates are reported, with $\Delta$Acc.
denoting the accuracy difference relative to the VLA-Cache-based pruning
baseline~\cite{xu2026vla} under the same setting. FLOPs, relative
FLOPs, CUDA latency, latency difference relative to vanilla OpenVLA, and
control frequency are also reported. \\The color annotations follow Table~\ref{tab:openvla-action-jnd-baselines}}
\vspace{-0.6em}
\label{tab:openvla-action-jnd-token-pruning}
\resizebox{\textwidth}{!}{%
\begin{tabular}{l|cccc|ccccccc}
\toprule[1.2pt]
\multirow{2}{*}{Method}
& \multicolumn{11}{c}{OpenVLA} \\
\cmidrule(lr){2-12}
& Spatial & Object & Goal & Long
& Acc.(\%)$\uparrow$
& $\Delta$Acc.$\uparrow$
& FLOPs(T)$\downarrow$
& Rel. FLOPs(\%)$\downarrow$
& Latency(ms)$\downarrow$
& $\Delta$Latency(ms)$\downarrow$
& Freq.(Hz)$\uparrow$ \\
\midrule

\rowcolor{gray!15}
\multicolumn{12}{c}{\emph{Upper Bound (Vanilla, 0\% Pruned)}} \\
Vanilla
& 79.80 & 61.40 & 72.60 & 51.80 & \textbf{66.40}
& -- & 1.985 & 100.00\% & 52.88 & -- & 18.91 \\

\midrule
\rowcolor{gray!15}
\multicolumn{12}{c}{\emph{Token Pruning Ratio = 25\%}} \\
FastV~\cite{chen2024image}
& 81.20 & 59.60 & 75.00 & 49.80 & \textbf{66.40}
& \textbf{-1.94} & 1.593 & 80.26\% & 40.46 & -12.43 & 24.72 \\
SparseVLM~\cite{zhang2024sparsevlm}
& 80.20 & 57.40 & 74.60 & 46.80 & \textbf{64.75}
& \textbf{-3.59} & 1.593 & 80.26\% & 41.40 & -11.49 & 24.16 \\
DivPrune~\cite{alvar2025divprune}
& \textcolor{negred}{77.40} & \textcolor{negred}{37.20} & \textcolor{negred}{74.40} & \textcolor{negred}{36.20} & \textcolor{negred}{\textbf{56.30}}
& \textcolor{negred}{\textbf{-12.04}} & \textcolor{posblue}{1.553} & \textcolor{posblue}{78.25\%} & \textcolor{posblue}{37.06} & \textcolor{posblue}{-15.82} & \textcolor{posblue}{26.98} \\
VLA-Cache~\cite{xu2026vla}
& 80.30 & \textcolor{posblue}{63.80} & 75.60 & \textcolor{posblue}{53.60} & \textcolor{posblue}{\textbf{68.34}}
& \textcolor{posblue}{\textbf{+0.00}} & 1.593 & 80.26\% & 40.53 & -12.36 & 24.67 \\
\rowcolor{actpink}
\textbf{Action-JND (Ours)} & \textcolor{posblue}{81.60} & 62.80 & \textcolor{posblue}{77.60} & 47.20 & \textbf{67.30} & \textbf{-1.04} & \textcolor{negred}{1.628} & \textcolor{negred}{81.99\%} & \textcolor{negred}{43.67} & \textcolor{negred}{-9.22} & \textcolor{negred}{22.90} \\

\midrule
\rowcolor{gray!15}
\multicolumn{12}{c}{\emph{Token Pruning Ratio = 50\%}} \\
FastV~\cite{chen2024image}
& 81.40 & 55.40 & 76.40 & 48.20 & \textbf{65.35}
& \textbf{-1.20} & 1.203 & 60.61\% & 36.58 & -16.31 & 27.34 \\
SparseVLM~\cite{zhang2024sparsevlm}
& 81.60 & 58.20 & 72.40 & 47.80 & \textbf{65.00}
& \textbf{-1.55} & 1.203 & 60.61\% & \textcolor{negred}{39.06} & \textcolor{negred}{-13.82} & \textcolor{negred}{25.60} \\
DivPrune~\cite{alvar2025divprune}
& \textcolor{negred}{46.60} & \textcolor{negred}{35.20} & \textcolor{negred}{66.20} & \textcolor{negred}{16.40} & \textcolor{negred}{\textbf{41.10}}
& \textcolor{negred}{\textbf{-25.45}} & \textcolor{posblue}{1.124} & \textcolor{posblue}{56.62\%} & \textcolor{posblue}{33.19} & \textcolor{posblue}{-19.70} & \textcolor{posblue}{30.13} \\
VLA-Cache~\cite{xu2026vla}
& 80.40 & 60.00 & 75.20 & \textcolor{posblue}{50.60} & \textbf{66.55}
& \textbf{+0.00} & 1.203 & 60.61\% & 34.59 & -18.30 & 28.91 \\
\rowcolor{actpink}
\textbf{Action-JND (Ours)} & \textcolor{posblue}{84.00} & \textcolor{posblue}{62.00} & \textcolor{posblue}{76.60} & 49.40 & \textcolor{posblue}{\textbf{68.00}} & \textcolor{posblue}{\textbf{+1.45}} & \textcolor{negred}{1.238} & \textcolor{negred}{62.34\%} & 37.74 & -15.15 & 26.50 \\

\midrule
\rowcolor{gray!15}
\multicolumn{12}{c}{\emph{Token Pruning Ratio = 75\%}} \\
FastV~\cite{chen2024image}
& \textcolor{posblue}{79.40} & 45.00 & 70.20 & 33.60 & \textbf{57.05}
& \textbf{-0.80} & 0.815 & 41.07\% & 32.05 & -20.84 & 31.20 \\
SparseVLM~\cite{zhang2024sparsevlm}
& 79.40 & 43.80 & 70.60 & 33.00 & \textbf{56.70}
& \textbf{-1.15} & 0.815 & 41.07\% & \textcolor{negred}{33.52} & \textcolor{negred}{-19.36} & \textcolor{negred}{29.83} \\
DivPrune~\cite{alvar2025divprune}
& \textcolor{negred}{6.60} & \textcolor{negred}{32.40} & \textcolor{negred}{39.60} & \textcolor{negred}{4.80} & \textcolor{negred}{\textbf{20.85}}
& \textcolor{negred}{\textbf{-37.00}} & \textcolor{posblue}{0.697} & \textcolor{posblue}{35.09\%} & \textcolor{posblue}{26.16} & \textcolor{posblue}{-26.72} & \textcolor{posblue}{38.22} \\
VLA-Cache~\cite{xu2026vla}
& 78.00 & 43.00 & 68.60 & 41.80 & \textbf{57.85}
& \textbf{+0.00} & 0.815 & 41.07\% & 31.02 & -21.87 & 32.24 \\
\rowcolor{actpink}
\textbf{Action-JND (Ours)} & 76.40 & \textcolor{posblue}{51.00} & \textcolor{posblue}{71.00} & \textcolor{posblue}{45.20} & \textcolor{posblue}{\textbf{60.90}} & \textcolor{posblue}{\textbf{+3.05}} & \textcolor{negred}{0.850} & \textcolor{negred}{42.80\%} & 31.05 & -21.84 & 32.21 \\

\midrule
\rowcolor{gray!15}
\multicolumn{12}{c}{\emph{Token Pruning Ratio = 87.5\%}} \\
FastV~\cite{chen2024image}
& 55.40 & 14.60 & 56.60 & 15.20 & \textbf{35.45}
& \textbf{+5.10} & 0.622 & 31.33\% & 28.95 & -23.94 & 34.55 \\
SparseVLM~\cite{zhang2024sparsevlm}
& 60.40 & \textcolor{negred}{7.40} & 56.00 & 13.80 & \textbf{34.40}
& \textbf{+4.05} & 0.622 & 31.33\% & 28.48 & -24.41 & 35.12 \\
DivPrune~\cite{alvar2025divprune}
& \textcolor{negred}{2.60} & 20.20 & \textcolor{negred}{19.00} & \textcolor{negred}{0.60} & \textcolor{negred}{\textbf{10.60}}
& \textcolor{negred}{\textbf{-19.75}} & \textcolor{posblue}{0.484} & \textcolor{posblue}{24.36\%} & \textcolor{posblue}{26.02} & \textcolor{posblue}{-26.86} & \textcolor{posblue}{38.43} \\
VLA-Cache~\cite{xu2026vla}
& 46.20 & 12.40 & 49.80 & 13.00 & \textbf{30.35}
& \textbf{+0.00} & 0.622 & 31.33\% & \textcolor{negred}{30.32} & \textcolor{negred}{-22.56} & \textcolor{negred}{32.98} \\
\rowcolor{actpink}
\textbf{Action-JND (Ours)} & \textcolor{posblue}{65.40} & \textcolor{posblue}{22.60} & \textcolor{posblue}{57.00} & \textcolor{posblue}{23.00} & \textcolor{posblue}{\textbf{42.00}} & \textcolor{posblue}{\textbf{+11.65}} & \textcolor{negred}{0.656} & \textcolor{negred}{33.06\%} & 30.16 & -22.72 & 33.15 \\

\bottomrule[1.2pt]
\end{tabular}%
}
\end{table*}

\vspace{-1em}
\subsection{Experimental Analysis}
Here we analyze the effectiveness of Action-JND under KV-cache reuse
and token pruning, focusing on task preservation, inference efficiency,
and qualitative compression behavior.

\subsubsection{Result Analysis on KV-Cache Reuse}
Tables~\ref{tab:openvla-action-jnd-baselines} and
\ref{tab:openvla-oft-action-jnd-baselines} show that Action-JND
provides more reliable KV-cache reuse on both OpenVLA and OpenVLA-OFT
backbones. Under both \texttt{soft} and
\texttt{strict\_guarded} reuse policies, Action-JND consistently achieves  
competitive or superior task performance across different reuse
ratios while maintaining computational costs comparable to those of existing methods. Notably, it achieves the highest  average accuracy under the \texttt{soft} policy on both backbones and generally exhibits a more favorable accuracy--efficiency trade-off under the \texttt{strict\_guarded} policy. At relatively low reuse ratios, performance differences among the compared methods are modest because only a conservative subset of visual representations is reused. As the reuse ratio increases, however, the
advantage of Action-JND becomes increasingly pronounced, whereas
methods based on manually designed compression criteria suffer rapid
performance degradation.

Existing methods mainly determine cache reuse using indirect compression cues such as attention, language relevance, visual redundancy, or proxy risk. In contrast, Action-JND directly estimates the action tolerance of each visual token and prioritizes stale-KV reuse for tokens whose representation changes are less likely to alter the predicted action, thereby better preserving action-critical information. At $60\%$ and $80\%$ reuse ratios, Action-JND improves the average accuracy over VLA-Cache~\cite{xu2026vla} by $23.70$ and $41.65$ percentage points on OpenVLA, and by $6.20$ and $10.10$ percentage points on OpenVLA-OFT, respectively. Under matched reuse ratios and comparable FLOPs, these gains demonstrate the effectiveness of the proposed action-conditioned tolerance criterion in identifying tokens suitable for stale-KV reuse. Action-JND also reduces CUDA latency and increases control frequency relative to vanilla inference, enabling more frequent action updates while better preserving task performance under aggressive KV-cache reuse.

\begin{figure*}[!t]
\centering
\vspace*{-2pt}
\includegraphics[width=0.95\linewidth]{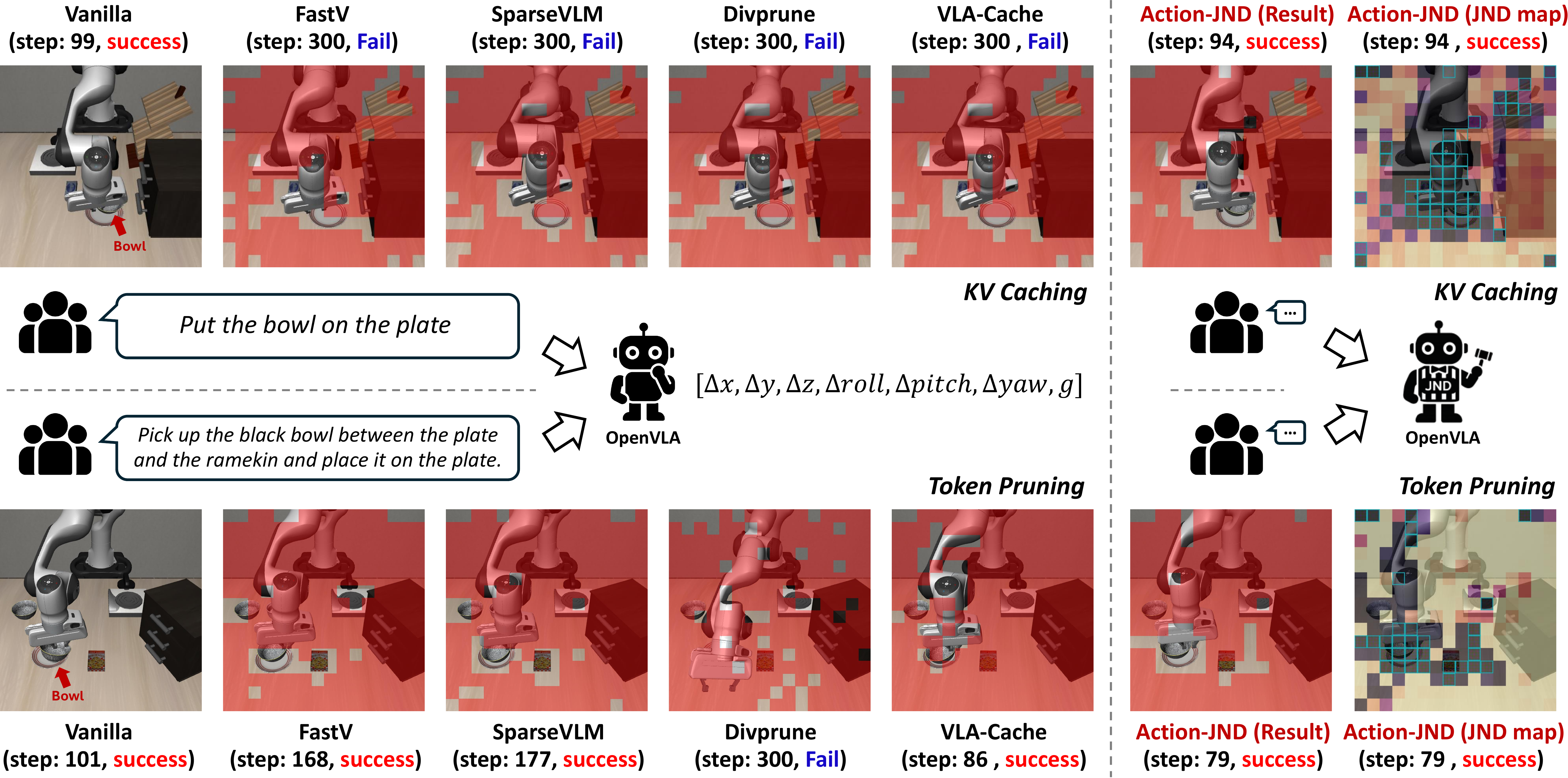}
\vspace*{-5pt}
\caption{\textbf{Qualitative visualization of different token-compression baselines and Action-JND-aware token compression.}
The upper and lower rows show representative results for KV-cache reuse
and token pruning on LIBERO, respectively. The \textcolor{red!80!black}{red} overlays indicate
visual-token regions selected for stale-KV reuse or physical pruning.
The last two columns show the Action-JND compression result and its
corresponding JND heatmap. Each heatmap visualizes the tolerance of individual visual tokens to feature perturbations. Color transitions from \textcolor{purple!70!black}{dark purple/black} to \textcolor{red!80!black}{red}, \textcolor{orange}{orange}, and \textcolor{yellow!80!orange}{bright yellow}, where \textcolor{yellow!80!orange}{brighter colors} indicate higher JND values and stronger action tolerance, making the corresponding tokens more suitable for reuse or pruning. In contrast, \textcolor{purple!70!black}{darker colors} indicate action-sensitive regions that should be preserved or recomputed. The \textcolor{cyan!70!blue}{blue} boxes in the JND heatmaps indicate the visual-token positions finally selected for stale-KV reuse or pruning. The step count denotes the number of control timesteps required for task execution, together with the final success or failure status.}
\label{fig:qualitative_results}
\vspace*{-12pt}
\end{figure*}

\subsubsection{Result Analysis on Token Pruning}
Table~\ref{tab:openvla-action-jnd-token-pruning} shows that Action-JND
provides more reliable visual-token pruning on OpenVLA. From $50\%$
pruning onward, Action-JND consistently achieves the highest average
accuracy among all compared schemes~\cite{alvar2025divprune,chen2024image,zhang2024sparsevlm,xu2026vla}, with its advantage becoming more
pronounced as the pruning ratio increases. Specifically, compared with
the VLA-Cache-based pruning baseline~\cite{xu2026vla}, Action-JND
improves the average accuracy by $1.45$, $3.05$, and $11.65$
percentage points at pruning ratios of $50\%$, $75\%$, and $87.5\%$,
respectively. It also consistently outperforms the other methods based on indirect pruning cues; even at
$87.5\%$ pruning, it exceeds the strongest competing method by
$6.55$ percentage points. These results suggest that compression
criteria based on indirect cues become increasingly unreliable under
aggressive pruning. Unlike KV-cache reuse, token pruning physically removes visual representations from all subsequent layers, making errors caused by removing action-sensitive tokens difficult to recover from in subsequent layers.

In contrast, Action-JND directly estimates the action tolerance of each
visual token and removes tokens whose absence is less likely to affect
the predicted action, thereby better preserving manipulation-critical
information. At the aggressive pruning ratio of $87.5\%$, Action-JND
maintains an average accuracy of $42.00\%$, compared with $30.35\%$
for the VLA-Cache-based pruning baseline and $35.45\%$ for the
strongest competing baseline. Meanwhile, it requires only $33.06\%$
of the FLOPs of the vanilla model, while reducing CUDA latency from
$52.88$ ms to $30.16$ ms and increasing control frequency from
$18.91$ to $33.15$ Hz. These results demonstrate that Action-JND
substantially reduces inference cost while retaining good
task performance, providing a favorable trade-off between action
prediction and inference efficiency.

\subsubsection{Qualitative Visualization}

Fig.~\ref{fig:qualitative_results} presents qualitative comparisons of
different token-compression baselines and Action-JND-aware token compression
for KV-cache reuse and token pruning under aggressive compression. These examples show that Action-JND better protects action-sensitive
information, enabling successful task execution in cases where several
competing methods fail or require more control steps.

\vspace*{-4pt}
\section{Conclusion}
In this paper, we put forward JND modeling toward embodied perception, and introduce Action-JND to
characterize the maximum visual-token variation tolerable to a VLA
policy without noticeably changing its action response. We further
instantiate this formulation with a lightweight token-wise JND estimator
and apply the learned action-tolerance scores to two representative token-compression paradigms, including KV-cache reuse and token pruning. Experiments on LIBERO with OpenVLA and OpenVLA-OFT demonstrate that
Action-JND enables more reliable compression, particularly under aggressive compression ratios, while improving inference efficiency for closed-loop robot control. In future work, we will extend embodied JND modeling to more diverse applications and tasks, including vision-language navigation with quadruped robots and coordinated multi-platform manipulation, and investigate its generalization across diverse embodiments and action spaces. 

{\footnotesize
\bibliographystyle{IEEEtran}
\bibliography{refs}

@article{zhang2023h2o,
  title={H2o: Heavy-hitter oracle for efficient generative inference of large language models},
  author={Zhang, Zhenyu and Sheng, Ying and Zhou, Tianyi and Chen, Tianlong and Zheng, Lianmin and Cai, Ruisi and Song, Zhao and Tian, Yuandong and R{\'e}, Christopher and Barrett, Clark and others},
  journal={Advances in Neural Information Processing Systems},
  volume={36},
  pages={34661--34710},
  year={2023}
}

@inproceedings{xiao2024efficient,
  title={Efficient streaming language models with attention sinks},
  author={Xiao, Guangxuan and Tian, Yuandong and Chen, Beidi and Han, Song and Lewis, Mike},
  booktitle={International Conference on Learning Representations},
  volume={2024},
  pages={21875--21895},
  year={2024}
}

@article{li2024snapkv,
  title={{SnapKV}: {LLM} knows what you are looking for before generation},
  author={Li, Yuhong and Huang, Yingbing and Yang, Bowen and Venkitesh, Bharat and Locatelli, Acyr and Ye, Hanchen and Cai, Tianle and Lewis, Patrick and Chen, Deming},
  journal={Advances in Neural Information Processing Systems},
  volume={37},
  pages={22947--22970},
  year={2024}
}

@article{cai2024pyramidkv,
  title={Pyramidkv: Dynamic kv cache compression based on pyramidal information funneling},
  author={Cai, Zefan and Zhang, Yichi and Gao, Bofei and Liu, Yuliang and Li, Yucheng and Liu, Tianyu and Lu, Keming and Xiong, Wayne and Dong, Yue and Hu, Junjie and others},
  journal={arXiv preprint arXiv:2406.02069},
  year={2024}
}

@article{bolya2022token,
  title={Token merging: Your vit but faster},
  author={Bolya, Daniel and Fu, Cheng-Yang and Dai, Xiaoliang and Zhang, Peizhao and Feichtenhofer, Christoph and Hoffman, Judy},
  journal={arXiv preprint arXiv:2210.09461},
  year={2022}
}

@inproceedings{chen2024image,
  title={An image is worth 1/2 tokens after layer 2: Plug-and-play inference acceleration for large vision-language models},
  author={Chen, Liang and Zhao, Haozhe and Liu, Tianyu and Bai, Shuai and Lin, Junyang and Zhou, Chang and Chang, Baobao},
  booktitle={European Conference on Computer Vision},
  pages={19--35},
  year={2024},
  organization={Springer}
}

@inproceedings{yang2025visionzip,
  title={Visionzip: Longer is better but not necessary in vision language models},
  author={Yang, Senqiao and Chen, Yukang and Tian, Zhuotao and Wang, Chengyao and Li, Jingyao and Yu, Bei and Jia, Jiaya},
  booktitle={Proceedings of the IEEE/CVF Conference on Computer Vision and Pattern Recognition},
  pages={19792--19802},
  year={2025}
}

@inproceedings{shang2025llava,
  title={Llava-prumerge: Adaptive token reduction for efficient large multimodal models},
  author={Shang, Yuzhang and Cai, Mu and Xu, Bingxin and Lee, Yong Jae and Yan, Yan},
  booktitle={Proceedings of the IEEE/CVF International Conference on Computer Vision},
  pages={22857--22867},
  year={2025}
}

@article{kim2024openvla,
  title={Openvla: An open-source vision-language-action model},
  author={Kim, Moo Jin and Pertsch, Karl and Karamcheti, Siddharth and Xiao, Ted and Balakrishna, Ashwin and Nair, Suraj and Rafailov, Rafael and Foster, Ethan and Lam, Grace and Sanketi, Pannag and others},
  journal={arXiv preprint arXiv:2406.09246},
  year={2024}
}

@article{kim2025fine,
  title={Fine-tuning vision-language-action models: Optimizing speed and success},
  author={Kim, Moo Jin and Finn, Chelsea and Liang, Percy},
  journal={arXiv preprint arXiv:2502.19645},
  year={2025}
}

@article{xu2026vla,
  title={{VLA-Cache}: Efficient vision-language-action manipulation via adaptive token caching},
  author={Xu, Siyu and Wang, Yunke and Xia, Chenghao and Zhu, Dihao and Huang, Tao and Xu, Chang},
  journal={Advances in Neural Information Processing Systems},
  volume={38},
  pages={164448--164473},
  year={2026}
}

@article{yang2026efficientvla,
  title={{EfficientVLA}: Training-free acceleration and compression for vision-language-action models},
  author={Yang, Yantai and Wang, Yuhao and Wen, Zichen and Zhongwei, Luo and Zou, Chang and Zhang, Zhipeng and Wen, Chuan and Zhang, Linfeng},
  journal={Advances in Neural Information Processing Systems},
  volume={38},
  pages={40891--40914},
  year={2026}
}

@article{liu2025vla,
  title={{VLA-Pruner}: Temporal-Aware Dual-Level Visual Token Pruning for Efficient Vision-Language-Action Inference},
  author={Liu, Ziyan and Chen, Yeqiu and Cai, Hongyi and Lin, Tao and Yang, Shuo and Liu, Zheng and Zhao, Bo},
  journal={arXiv preprint arXiv:2511.16449},
  year={2025}
}

@article{sullivan1998rate,
  title={Rate-distortion optimization for video compression},
  author={Sullivan, Gary J and Wiegand, Thomas},
  journal={IEEE signal processing magazine},
  volume={15},
  number={6},
  pages={74--90},
  year={1998},
  publisher={IEEE}
}

@article{chou1995perceptually,
  title={A perceptually tuned subband image coder based on the measure of just-noticeable-distortion profile},
  author={Chou, Chun-Hsien and Li, Yun-Chin},
  journal={IEEE Transactions on circuits and systems for video technology},
  volume={5},
  number={6},
  pages={467--476},
  year={1995},
  publisher={IEEE}
}

@article{jayant2002signal,
  title={Signal compression based on models of human perception},
  author={Jayant, Nikil and Johnston, James and Safranek, Robert},
  journal={Proceedings of the IEEE},
  volume={81},
  number={10},
  pages={1385--1422},
  year={2002},
  publisher={IEEE}
}

@article{lin2021progress,
  title={Progress and opportunities in modelling just-noticeable difference (JND) for multimedia},
  author={Lin, Weisi and Ghinea, Gheorghita},
  journal={IEEE transactions on multimedia},
  volume={24},
  pages={3706--3721},
  year={2021},
  publisher={IEEE}
}

@article{chen2025just,
  title={Just noticeable difference for large multimodal models},
  author={Chen, Zijian and Tian, Yuan and Sun, Yuze and Sun, Wei and Zhang, Zicheng and Lin, Weisi and Zhai, Guangtao and Zhang, Wenjun},
  journal={arXiv preprint arXiv:2507.00490},
  year={2025}
}

@inproceedings{zhao2026just,
  title={Just Noticeable Difference Modeling for Deep Visual Features},
  author={Zhao, Rui and Li, Wenrui and Zhu, Lin and Zheng, Yajing and Lin, Weisi},
  booktitle={International Conference on Machine Learning},
  year={2026}
}

@inproceedings{zitkovich2023rt,
  title={{RT}-2: Vision-language-action models transfer web knowledge to robotic control},
  author={Zitkovich, Brianna and Yu, Tianhe and Xu, Sichun and Xu, Peng and Xiao, Ted and Xia, Fei and Wu, Jialin and Wohlhart, Paul and Welker, Stefan and Wahid, Ayzaan and others},
  booktitle={Conference on Robot Learning},
  pages={2165--2183},
  year={2023},
  organization={PMLR}
}

@article{black2024pi_0,
  title={$\pi_0$: A Vision-Language-Action Flow Model for General Robot Control},
  author={Black, Kevin and Brown, Noah and Driess, Danny and Esmail, Adnan and Equi, Michael and Finn, Chelsea and Fusai, Niccolo and Groom, Lachy and Hausman, Karol and Ichter, Brian and others},
  journal={arXiv preprint arXiv:2410.24164},
  year={2024}
}

@article{team2024octo,
  title={Octo: An open-source generalist robot policy},
  author={Team, Octo Model and Ghosh, Dibya and Walke, Homer and Pertsch, Karl and Black, Kevin and Mees, Oier and Dasari, Sudeep and Hejna, Joey and Kreiman, Tobias and Xu, Charles and others},
  journal={arXiv preprint arXiv:2405.12213},
  year={2024}
}

@article{zhang2024sparsevlm,
  title={{SparseVLM}: Visual token sparsification for efficient vision-language model inference},
  author={Zhang, Yuan and Fan, Chun-Kai and Ma, Junpeng and Zheng, Wenzhao and Huang, Tao and Cheng, Kuan and Gudovskiy, Denis and Okuno, Tomoyuki and Nakata, Yohei and Keutzer, Kurt and others},
  journal={arXiv preprint arXiv:2410.04417},
  year={2024}
}

@inproceedings{alvar2025divprune,
  title={{DivPrune}: Diversity-based visual token pruning for large multimodal models},
  author={Alvar, Saeed Ranjbar and Singh, Gursimran and Akbari, Mohammad and Zhang, Yong},
  booktitle={Proceedings of the Computer Vision and Pattern Recognition Conference},
  pages={9392--9401},
  year={2025}
}

@article{li2025sp,
  title={{SP-VLA}: A joint model scheduling and token pruning approach for vla model acceleration},
  author={Li, Ye and Meng, Yuan and Sun, Zewen and Ji, Kangye and Tang, Chen and Fan, Jiajun and Ma, Xinzhu and Xia, Shutao and Wang, Zhi and Zhu, Wenwu},
  journal={arXiv preprint arXiv:2506.12723},
  year={2025}
}

@article{wang2025specprune,
  title={{SpecPrune‑VLA}: Accelerating vision-language-action models via action-aware self-speculative pruning},
  author={Wang, Hanzhen and Xu, Jiaming and Xiang, Yushun and Pan, Jiayi and Zhou, Yongkang and Li, Yong-Lu and Dai, Guohao},
  journal={arXiv preprint arXiv:2509.05614},
  year={2025}
}

@article{jiang2025better,
  title={The better you learn, the smarter you prune: Towards efficient vision-language-action models via differentiable token pruning},
  author={Jiang, Titong and Jiang, Xuefeng and Ma, Yuan and Wen, Xin and Li, Bailin and Zhan, Kun and Jia, Peng and Liu, Yahui and Sun, Sheng and Lang, Xianpeng},
  journal={arXiv preprint arXiv:2509.12594},
  year={2025}
}

@article{pei2025action,
  title={Action-aware dynamic pruning for efficient vision-language-action manipulation},
  author={Pei, Xiaohuan and Chen, Yuxing and Xu, Siyu and Wang, Yunke and Shi, Yuheng and Xu, Chang},
  journal={arXiv preprint arXiv:2509.22093},
  year={2025}
}

@article{fang2025sqap,
  title={{SQAP-VLA}: A synergistic quantization-aware pruning framework for high-performance vision-language-action models},
  author={Fang, Hengyu and Liu, Yijiang and Du, Yuan and Du, Li and Yang, Huanrui},
  journal={arXiv preprint arXiv:2509.09090},
  year={2025}
}

@article{jin2021just,
  title={Just noticeable difference for deep machine vision},
  author={Jin, Jian and Zhang, Xingxing and Fu, Xin and Zhang, Huan and Lin, Weisi and Lou, Jian and Zhao, Yao},
  journal={IEEE Transactions on Circuits and Systems for Video Technology},
  volume={32},
  number={6},
  pages={3452--3461},
  year={2021},
  publisher={IEEE}
}

@article{zhang2021just,
  title={Just recognizable distortion for machine vision oriented image and video coding},
  author={Zhang, Qi and Wang, Shanshe and Zhang, Xinfeng and Ma, Siwei and Gao, Wen},
  journal={International Journal of Computer Vision},
  volume={129},
  number={10},
  pages={2889--2906},
  year={2021},
  publisher={Springer}
}

@article{zhang2024perceptual,
  title={Perceptual video coding for machines via satisfied machine ratio modeling},
  author={Zhang, Qi and Wang, Shanshe and Zhang, Xinfeng and Jia, Chuanmin and Wang, Zhao and Ma, Siwei and Gao, Wen},
  journal={IEEE Transactions on Pattern Analysis and Machine Intelligence},
  volume={46},
  number={12},
  pages={7651--7668},
  year={2024},
  publisher={IEEE}
}

@article{lin2022large,
  title={Large-scale crowdsourced subjective assessment of picturewise just noticeable difference},
  author={Lin, Hanhe and Chen, Guangan and Jenadeleh, Mohsen and Hosu, Vlad and Reips, Ulf-Dietrich and Hamzaoui, Raouf and Saupe, Dietmar},
  journal={IEEE transactions on circuits and systems for video technology},
  volume={32},
  number={9},
  pages={5859--5873},
  year={2022},
  publisher={IEEE}
}

@inproceedings{chen2023localization,
  title={Localization of just noticeable difference for image compression},
  author={Chen, Guangan and Lin, Hanhe and Wiedemann, Oliver and Saupe, Dietmar},
  booktitle={2023 15th International Conference on Quality of Multimedia Experience (QoMEX)},
  pages={61--66},
  year={2023},
  organization={IEEE}
}

@article{jiang2022toward,
  title={Toward top-down just noticeable difference estimation of natural images},
  author={Jiang, Qiuping and Liu, Zhentao and Wang, Shiqi and Shao, Feng and Lin, Weisi},
  journal={IEEE Transactions on Image Processing},
  volume={31},
  pages={3697--3712},
  year={2022},
  publisher={IEEE}
}

@inproceedings{wang2024metajnd,
  title={{MetaJND}: A Meta-Learning Approach for Just Noticeable Difference Estimation.},
  author={Wang, Miaohui and Zhu, Yukuan and Zhang, Rong and Xie, Wuyuan},
  booktitle={IJCAI},
  pages={3151--3159},
  year={2024}
}

@article{liu2023libero,
  title={Libero: Benchmarking knowledge transfer for lifelong robot learning},
  author={Liu, Bo and Zhu, Yifeng and Gao, Chongkai and Feng, Yihao and Liu, Qiang and Zhu, Yuke and Stone, Peter},
  journal={Advances in Neural Information Processing Systems},
  volume={36},
  pages={44776--44791},
  year={2023}
}

@article{gao2024dmofc,
  title={{DMOFC}: Discrimination metric-optimized feature compression},
  author={Gao, Changsheng and Jiang, Yiheng and Li, Li and Liu, Dong and Wu, Feng},
  journal={arXiv preprint arXiv:2405.04044},
  year={2024}
}

@inproceedings{gao2025feature,
  title={Feature coding in the era of large models: Dataset, test conditions, and benchmark},
  author={Gao, Changsheng and Ma, Yifan and Chen, Qiaoxi and Xu, Yenan and Liu, Dong and Lin, Weisi},
  booktitle={Proceedings of the IEEE/CVF International Conference on Computer Vision},
  pages={1068--1077},
  year={2025}
}

@inproceedings{gao2025dt,
  title={{DT-UFC}: Universal Large Model Feature Coding via Peaky-to-Balanced Distribution Transformation},
  author={Gao, Changsheng and Liu, Zijie and Li, Li and Liu, Dong and Sun, Xiaoyan and Lin, Weisi},
  booktitle={Proceedings of the 33rd ACM International Conference on Multimedia},
  pages={5198--5207},
  year={2025}
}

@inproceedings{gao2025compressed,
  title={Compressed feature quality assessment: Dataset and baselines},
  author={Gao, Changsheng and Zhou, Wei and Lin, Guosheng and Lin, Weisi},
  booktitle={Proceedings of the 33rd ACM International Conference on Multimedia},
  pages={13450--13456},
  year={2025}
}

@article{gao2021towards,
  title={Towards task-generic image compression: A study of semantics-oriented metrics},
  author={Gao, Changsheng and Liu, Dong and Li, Li and Wu, Feng},
  journal={IEEE Transactions on Multimedia},
  volume={25},
  pages={721--735},
  year={2021},
  publisher={IEEE}
}

@article{huang2013feature,
  title={Feature coding in image classification: A comprehensive study},
  author={Huang, Yongzhen and Wu, Zifeng and Wang, Liang and Tan, Tieniu},
  journal={IEEE transactions on pattern analysis and machine intelligence},
  volume={36},
  number={3},
  pages={493--506},
  year={2013},
  publisher={IEEE}
}

@article{wu2013just,
  title={Just noticeable difference estimation for images with free-energy principle},
  author={Wu, Jinjian and Shi, Guangming and Lin, Weisi and Liu, Anmin and Qi, Fei},
  journal={IEEE Transactions on Multimedia},
  volume={15},
  number={7},
  pages={1705--1710},
  year={2013},
  publisher={IEEE}
}

@article{shen2025transferring,
  title={Transferring From Distortion to Perception-Oriented Optimization: Just-Noticeable-Distortion-Based Domain Adaptation},
  author={Shen, Xuelin and Ou, Haoqiao and Ni, Zhangkai and Yang, Wenhan and Wang, Shiqi and Kwong, Sam},
  journal={IEEE Transactions on Multimedia},
  year={2025},
  publisher={IEEE}
}

@article{li2025embodied,
  title={Embodied Image Compression},
  author={Li, Chunyi and Qing, Rui and Zhang, Jianbo and Tian, Yuan and Zhu, Xiangyang and Zhang, Zicheng and Liu, Xiaohong and Lin, Weisi and Zhai, Guangtao},
  journal={arXiv preprint arXiv:2512.11612},
  year={2025}
}

@inproceedings{li2026image,
  title={Image quality assessment for embodied ai},
  author={Li, Chunyi and Xiao, Jiahao and Zhang, Jianbo and Wen, Farong and Zhang, Zicheng and Tian, Yuan and Zhu, Xiangyang and Liu, Xiaohong and Cheng, Zhengxue and Lin, Weisi and others},
  booktitle={International Conference on Learning Representations},
  volume={2026},
  pages={39960--39982},
  year={2026}
}

@article{li2025ustc,
  title={{USTC-TD}: A test dataset and benchmark for image and video coding in 2020s},
  author={Li, Zhuoyuan and Liao, Junqi and Tang, Chuanbo and Zhang, Haotian and Li, Yuqi and Bian, Yifan and Sheng, Xihua and Feng, Xinmin and Li, Yao and Gao, Changsheng and others},
  journal={IEEE Transactions on Multimedia},
  year={2025},
  publisher={IEEE}
}

@article{bross2021overview,
  title={Overview of the versatile video coding ({VVC}) standard and its applications},
  author={Bross, Benjamin and Wang, Ye-Kui and Ye, Yan and Liu, Shan and Chen, Jianle and Sullivan, Gary J and Ohm, Jens-Rainer},
  journal={IEEE Transactions on Circuits and Systems for Video Technology},
  volume={31},
  number={10},
  pages={3736--3764},
  year={2021},
  publisher={IEEE}
}

@article{sullivan2012overview,
  title={Overview of the high efficiency video coding (HEVC) standard},
  author={Sullivan, Gary J and Ohm, Jens-Rainer and Han, Woo-Jin and Wiegand, Thomas},
  journal={IEEE Transactions on circuits and systems for video technology},
  volume={22},
  number={12},
  pages={1649--1668},
  year={2012},
  publisher={IEEE}
}
}

\begin{IEEEbiography}
[{\includegraphics[width=1in,height=1.25in,clip,keepaspectratio]{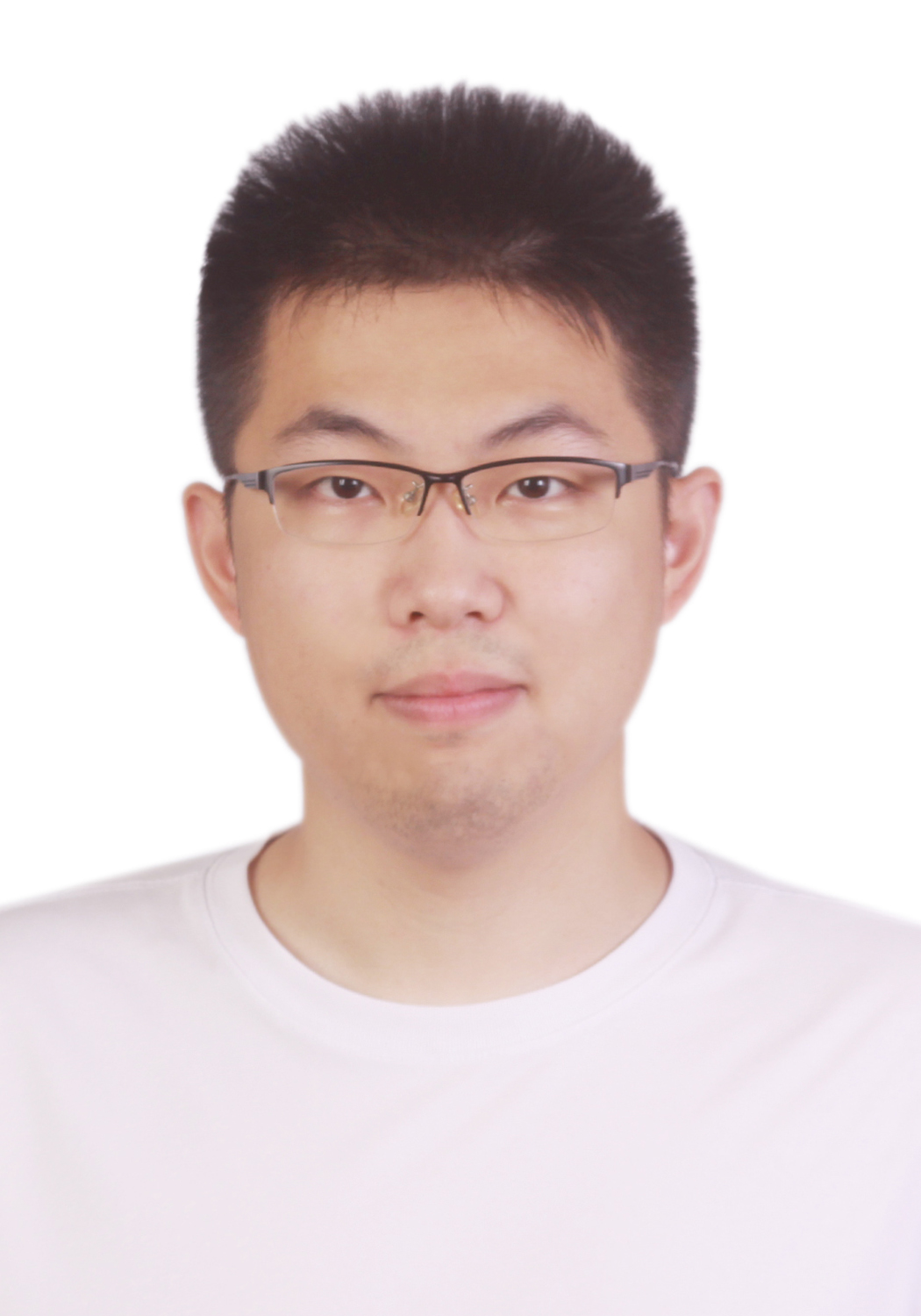}}] 
{Zhuoyuan Li}  
(Member, IEEE) received the B.S. degree in communication engineering from Southwest Jiaotong University, China, in 2020, and the Ph.D. degree in electronic engineering and information science from the University of Science and Technology of China (USTC), China, in 2025. He is currently a Postdoctoral Researcher at The Hong Kong Polytechnic University (PolyU), supervised by Kenneth Kin-man Lam.
His research interests include image and video coding. He has published over 30 papers in international journals and conferences. He won several technical challenges at ICIP 2024, MMSP 2024, and VCIP 2025. He received the PRCV 2025 Grand Challenge Outstanding Exploration Award and the NeurIPS 2025 Top Reviewer Award. He has submitted several technical proposals to ISO/IEC and AVS standardization activities.
\end{IEEEbiography}

\begin{IEEEbiography}
[{\includegraphics[width=1in,height=1.25in,clip,keepaspectratio]{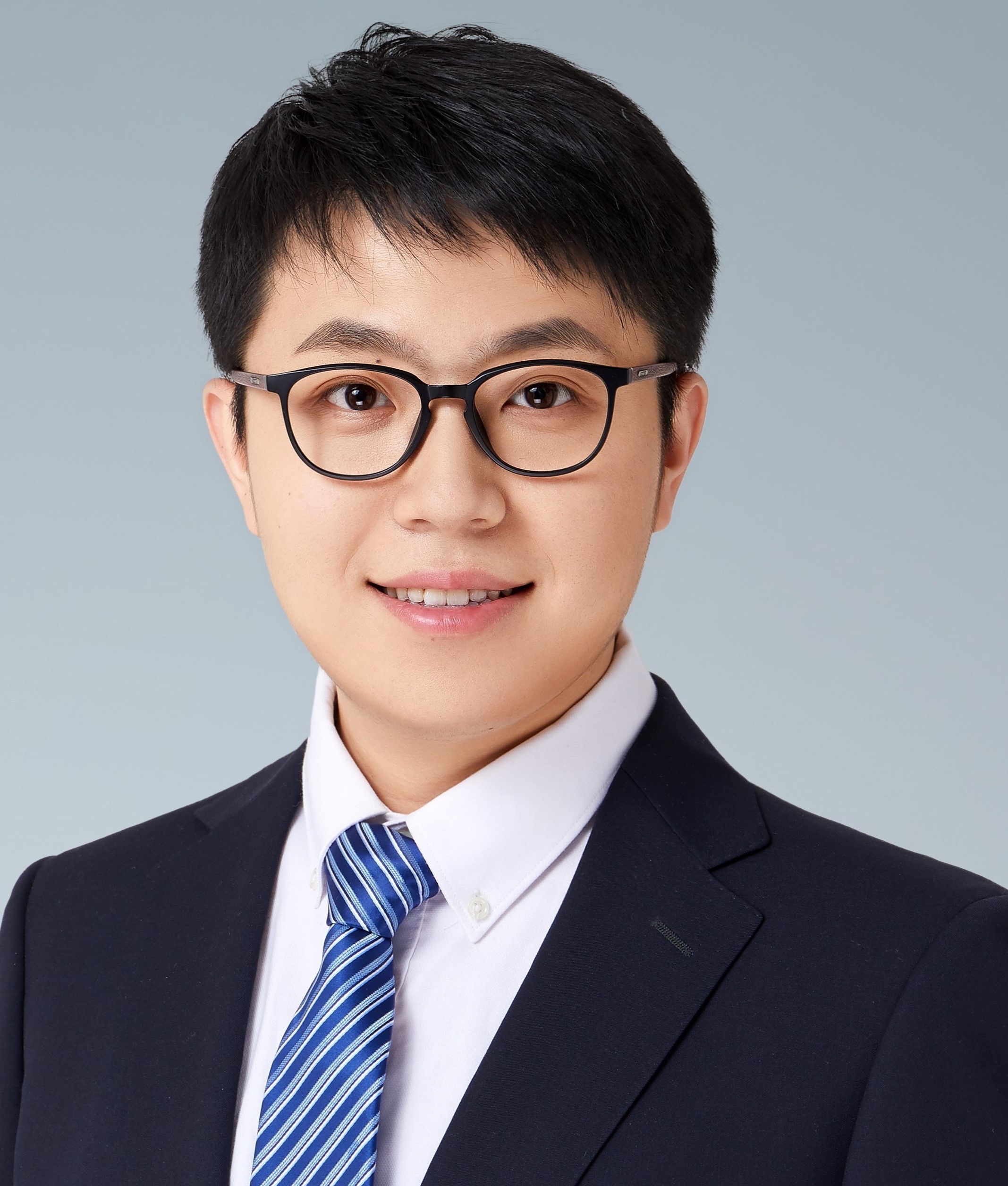}}] 
{Rui Zhao} 
(Member, IEEE) received the B.E. degree in Communication Engineering from Tianjin University, Tianjin, China, in 2020, and the Ph.D. degree in Computer Science from Peking University, Beijing, China, in 2025.
He is currently a postdoctoral research fellow with Nanyang Technological University, Singapore. His research interests include computational photography, computer vision, and image processing, particularly for topics related to motion estimation and neuromorphic cameras.
\end{IEEEbiography}

\begin{IEEEbiography}[{\includegraphics[width=1in,height=1.25in, clip,keepaspectratio]{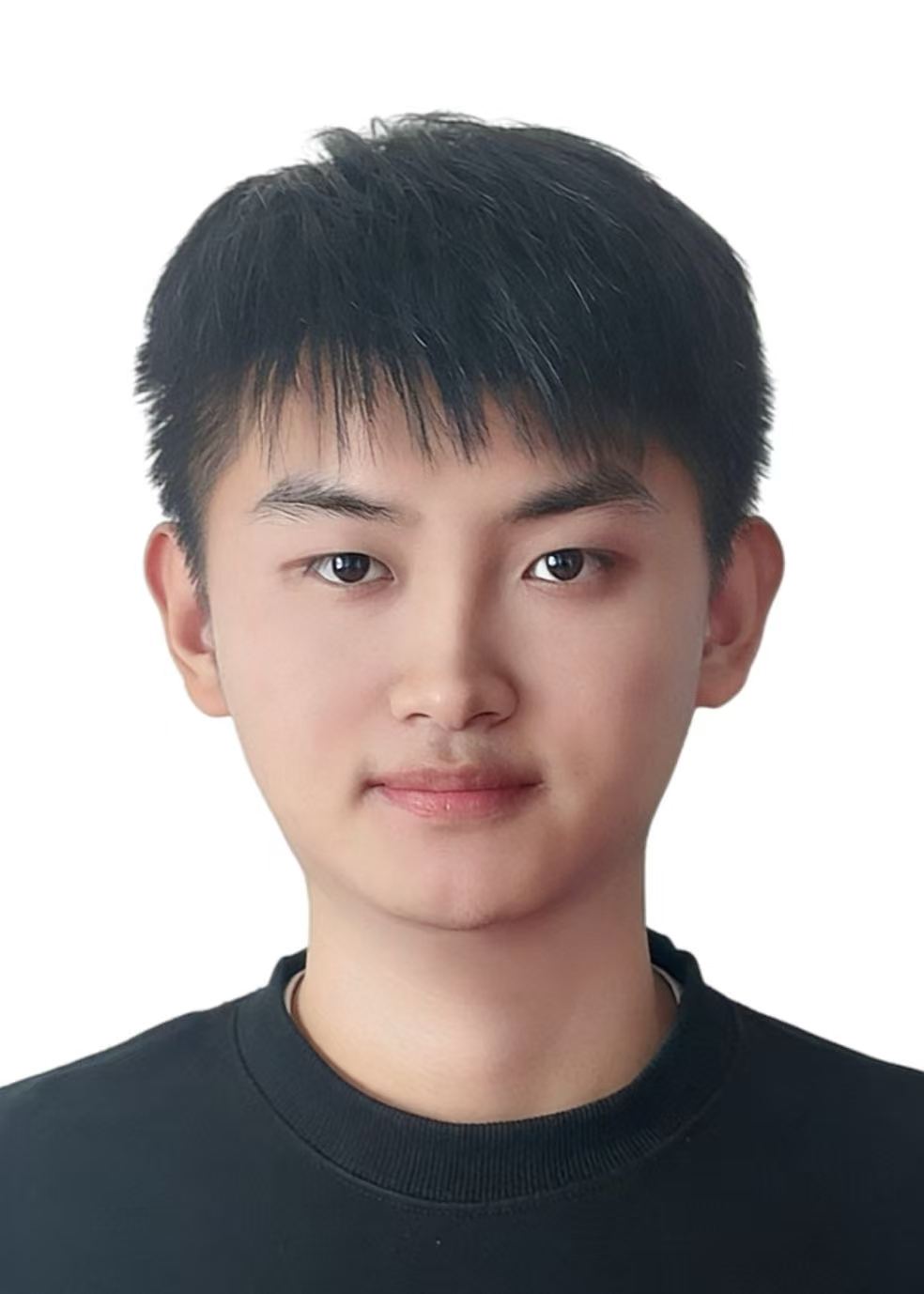}}]
{Jin Wang} received the M.S. degree in electronic information from the University of Science and Technology of China (USTC), Hefei, China, in 2024. His research interests include computational photography, depth sensing, and vision-language-action models for embodied intelligence.
\end{IEEEbiography}

\begin{IEEEbiography}[{\includegraphics[width=1in,height=1.25in, clip,keepaspectratio]{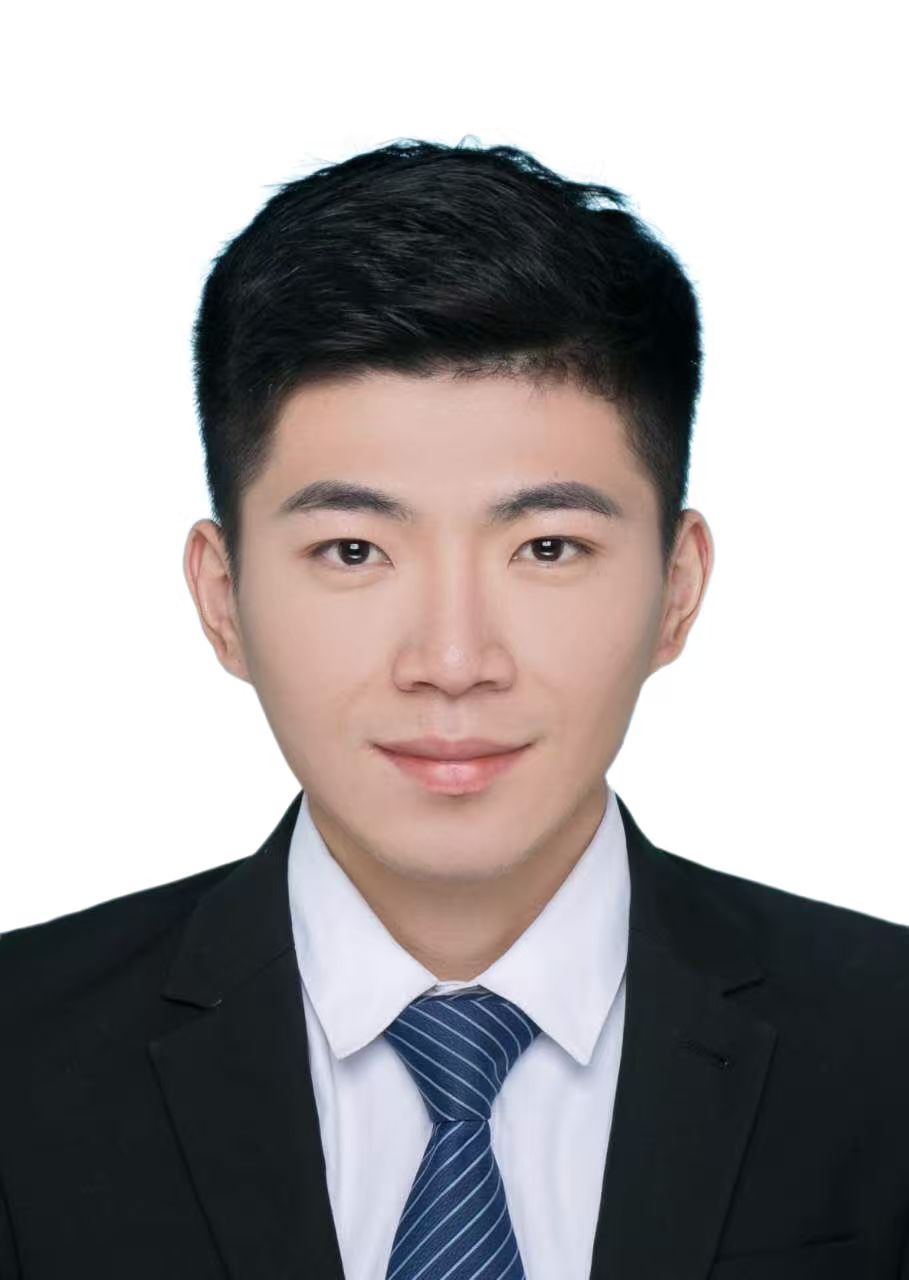}}]
{Hanwei Zhu} (Member, IEEE) received  Ph.D. degree in computer science from the City University of Hong Kong, Hong Kong, in 2025.  He is currently a research scientist with the Alibaba-NTU Global e-Sustainability CorpLab (ANGEL) at Nanyang Technological University. His research interests include perceptual image processing, computational vision, and computational photography.
\end{IEEEbiography}

\begin{IEEEbiography}[{\includegraphics[width=1in,height=1.25in, clip,keepaspectratio]{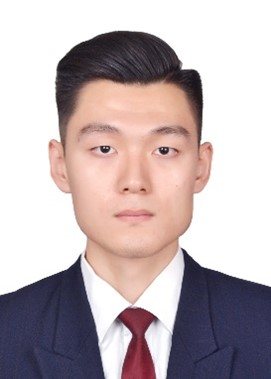}}]
{Cong Zhang} (Member, IEEE) received the Ph.D. degree from the Department of Electrical and Electronic Engineering, The Hong Kong Polytechnic University, the B.E. degree and the M.E. degree from the School of Artificial Intelligence, Optics and Electronics (iOPEN), Northwestern Polytechnical University. He was a
Postdoctoral Fellow at The Hong Kong Polytechnic
University and The Chinese University of Hong
Kong. Currently, he is a Professor with the School
of Control Science and Engineering, Shandong University.\end{IEEEbiography}

\begin{IEEEbiography}
[{\includegraphics[width=1in,height=1.25in,clip,keepaspectratio]{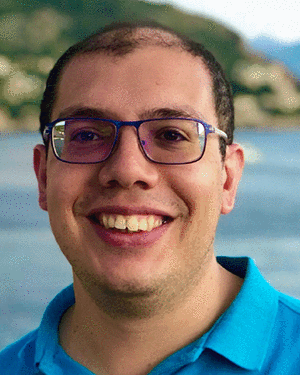}}] 
{Giuseppe Valenzise} (Senior Member, IEEE) is a CNRS researcher at the Laboratoire des Signaux et Systèmes, CentraleSupélec, Université Paris-Saclay, France, where he leads the Multimedia and Networking team. He is Editor-in-Chief of the Springer EURASIP Journal on Image and Video Processing. He received his PhD from Politecnico di Milano, Italy. His research covers image and video processing, with a focus on compression (traditional and learning-based), light field and point cloud coding, quality assessment, high dynamic range imaging, and machine learning for visual analysis. He has co-authored over 100 publications in these areas and received the EURASIP Early Career Award in 2018. Giuseppe was General Chair of ICME 2025 and regularly serves on organizing committees of flagship conferences such as ICIP. He has been Associate Editor for IEEE TCSVT, IEEE TIP, and Signal Processing: Image Communication. He is currently Chair of the IEEE SPS Multimedia Signal Processing Technical Committee and a member of the IEEE CAS Multimedia Systems and Applications Technical Committee.
\end{IEEEbiography}

\begin{IEEEbiography}[{\includegraphics[width=1in,height=1.25in,clip,keepaspectratio]{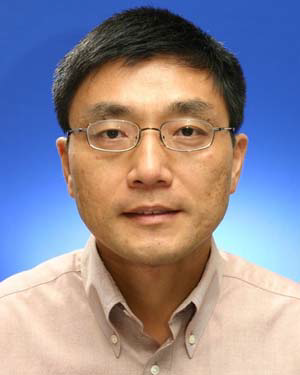}}]
{Weisi Lin} (Fellow, IEEE) is a President’s Chair Professor in Computer Science, Nanyang Technological University, Singapore. His research interests include intelligent image processing, perceptual signal modeling, video compression, and multimedia communication. He is a Chartered Engineer and a fellow of  IET. He serves as a General Co-Chair for IEEE ICME 2025 and Lead General Chair for IEEE ICIP 2027, and was a Technical Program Chair of the IEEE ICME 2013, PCM 2012, QoMEX 2014/2026 and IEEE VCIP 2017. He has been a Keynote/Invited/Panelist/Tutorial Speaker at over 50 international conferences, and was a Distinguished Lecturer of the IEEE Circuits and Systems Society from 2016 to 2017 and the AsiaPacific Signal and Information Processing Association (APSIPA) from 2012 to 2013. He has been an Associate Editor of the IEEE Trans. Neural Networks and Learning Syst., the IEEE Trans. Image Process., the IEEE Trans. Circuits Syst. Video Technol., the IEEE Trans. Multimedia, and the IEEE Signal Process. Lett., as well as a Senior Editor for IEEE J. Selected Topics in Sig. Process. He has been awarded the Research Award 2023, College of Engineering, NTU, and is a Highly Cited Researcher since 2019 (awarded by Clarivate Analytics).
\end{IEEEbiography}

\begin{IEEEbiography}[{\includegraphics[width=1in,height=1.25in,clip,keepaspectratio]{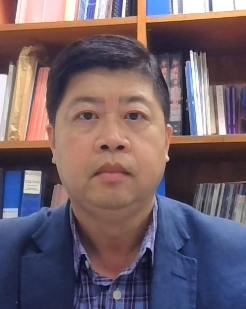}}]
{Kin-Man Lam} (Senior Member, IEEE) received his Associateship in Electronic Engineering with distinction from Hong Kong Polytechnic in 1986, his M.Sc. degree in Communication Engineering from Imperial College in 1987, and his Ph.D. degree from the University of Sydney in 1996. He joined the Department of Electronic and Information Engineering, The Hong Kong Polytechnic University as an Assistant Professor in October 1996. He became an Associate Professor in 1999 and has been a Professor since 2010. He served as Chair of the IEEE Hong Kong Chapter of Signal Processing from 2006 to 2008. He was Technical Co-Chair of ISPACS 2005, PCM 2010, and IEEE VCIP 2020, and General Co-Chair of ICSPCC 2012, APSIPA ASC 2015, and IEEE ICME 2017. He also served as Director-Student Services (2012-2014), Director-Membership Services (2015-2017), and VP-Membership (2023-2025) for the IEEE SPS. Additionally, he held the roles of VP-Member Relations and Development (2014-2017) and VP-Publications (2017-2021) for APSIPA. He was an Associate Editor for IEEE Trans. on Image Processing (2009-2014) and Digital Signal Processing (2014-2018), an Editor for HKIE Transactions (2013-2018), and an Area Editor for the IEEE Signal Processing Magazine (2015-2017). He currently serves as a Senior Editorial Board Member of APSIPA Trans. on Signal and Information Processing and an Associate editor of the EURASIP International Journal on Image and Video Processing. His current research interests include image and video processing, computer vision, and human face analysis and recognition.
\end{IEEEbiography}

\end{document}